\documentclass[times,preprint,referee,twocolumn,authoryear]{elsarticle_modified}

\usepackage{ycviu_modified}
\usepackage{framed,multirow}

\usepackage{amssymb}
\usepackage{latexsym}
\usepackage{amsmath}
\usepackage{amssymb}
\usepackage{pifont}

\usepackage[hyphens]{url}
\usepackage{xcolor}
\definecolor{newcolor}{rgb}{.8,.349,.1}
\usepackage[breaklinks=true,bookmarks=false]{hyperref}
\hypersetup{
	colorlinks=true,
	linkcolor=newcolor,
	filecolor=newcolor,      
	urlcolor=cyan
}

\journal{Computer Vision and Image Understanding}

\begin{document}
	
\ifpreprint
\setcounter{page}{1}
\else
\setcounter{page}{1}
\fi
	
\twocolumn[{\begin{frontmatter}
			
\title{A review of 3D human pose estimation algorithms for markerless motion capture}
\vspace*{1.8cm}
\author[1]{Yann \snm{Desmarais}}
\author[2]{Denis \snm{Mottet}}
\author[1]{Pierre \snm{Slangen}}
\author[1]{Philippe \snm{Montesinos}\corref{cor1}}
\cortext[cor1]{Corresponding author: 
Tel.: +33-(0)4-34-24-62-95}
\ead{philippe.montesinos@mines-ales.fr}

\address[1]{ EuroMov Digital Health in Motion, Univ Montpellier, IMT Mines Ales, 30100 Ales, France}
\address[2]{ EuroMov Digital Health in Motion, Univ Montpellier, IMT Mines Ales, 34090 Montpellier, France}
			
\textbf{This article is under consideration at Computer Vision and Image Understanding.}

\begin{abstract}
Human pose estimation is a very active research field, stimulated by its important applications in robotics, entertainment or health and sports sciences, among others. Advances in convolutional networks triggered noticeable improvements in 2D pose estimation, leading modern 3D markerless motion capture techniques to an average error per joint of 20 mm. However, with the proliferation of methods, it is becoming increasingly difficult to make an informed choice.
Here, we review the leading human pose estimation methods of the past five years, focusing on metrics, benchmarks and method structures. We propose a taxonomy based on accuracy, speed and robustness that we use to classify de methods and derive directions for future research.
\end{abstract}


\end{frontmatter}}]

\section{Introduction}

Human Pose Estimation is the extraction of body configurations in images or videos. Typically, it is the inference of joint coordinates and the reconstruction of a human skeletal representation. In the last few years, 2D pose estimation reached detection rate above 90\% on all different human joints \cite{newell_stacked_2016}. This progress has been possible in great part because of the success of convolutional neural networks (CNN) and the appearance of accessible large scale datasets (\cite{sigal_humaneva:_2010}; \cite{ionescu_human3.6m:_2014}). However, it is only recently that these new architectures have been deployed to solve a similar problem in 3D. The challenge for these new 3D markerless pose estimation methods is to be competitive against classical techniques and marker-based motion capture systems. The ultimate goal would be a complete and accurate 3D reconstruction of an individual's motion from simple monocular images with tolerance to severe occlusion. As this ideal is unrealistic, results on similar tasks seem to indicate that it is possible to reach some of these conditions even if all are not fulfilled.\\

Traditionally, commercial motion capture systems track small reflective markers placed on the surface of subjects. While precise, traditional motion capture is heavily constrained by complex sensors and acquisition environment. In 2010, Microsoft released the Kinect sensor that captured human pose from RGB and depth images. However, it was mostly aimed at entertainment usage and was not suitable for outdoor acquisition.\\

However, pose and person detection algorithms based only on image features have existed since a long time. The former paradigm was the fitting of human defined features to complex part-based human models (representations of the silhouette with cylinders, stick-figures, meshes, cones or boxes). While some modern techniques still use that approach, the feature extraction part of the process is now realized using convolutional neural networks.\\
\subsection{Other Surveys}

\begin{figure}[ht]
	\begin{center}
		\includegraphics[width=0.3\textwidth]{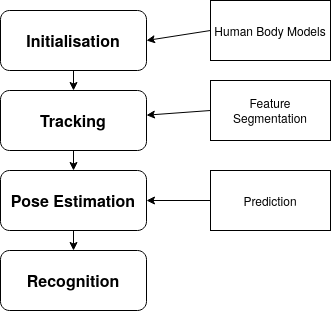}
	\end{center}
	\caption{Classical motion capture system }
	\label{mocap}
\end{figure}

\cite{bray_markerless_2000} reviews optical markerless methods with a taxonomy based on commonly performed subtasks: Initialization, Tracking, Pose Estimation and Recognition (Fig.\ref{mocap}). With this classification, they describe the different ways to extract visual features for the pose estimation process and then how to track them between frames.\\

The survey of \cite{moeslund_survey_2001} explores twenty years of vision-based human pose estimation techniques from 1980 to the early 2000s, including marker-based and markerless methods. At that time, many assumptions were taken to facilitate the process of extracting the human silhouette: most methods functioned indoor with non-shifting lighting and nearly half of them used uniform static backgrounds. This survey provides a good history of the different families of pose estimation. The authors also carefully describe the different degrees of performances needed for applications using human pose estimation and how to quantify them. However, their functionality-based taxonomy is no longer adapted for today's techniques, because modern methods mostly do not use an initialization step and perform pose estimation and tracking at the same time. Finally, action recognition is nowadays a computer vision task with its own community.\\

\cite{sarafinos_pdf_2016} wrote a review of more recent methods with an emphasis on input data. They suggest a taxonomy that divides pose estimation between monocular versus multi-view techniques and between static image versus video inputs. This distinction is still present in our review, but our analysis is more oriented towards the family of method employed (learning, model-based etc.). They also evaluate part-based and learning-based methods on modern benchmarks. They also propose a synthetic dataset (SynPose300) to evaluate robustness (erroneous initialization, viewing distance, difficult poses or actions). The focus is made on methods from 2008 until 2016. We chose to start our review from 2017, to present recent evolutions in 3D pose estimation.\\  

\cite{colyer_review_2018} examined the historical methods used in biomechanical studies of human motion. The authors insist that training and validating markerless methods with marker-based ones is not completely correct, because the reflecting markers can modify the results (helping or degrading the results otherwise obtainable in an "in-the-wild" context). They also state that marker-based methods are not providing the accuracy of measurement to characterize real human movement for some sport science and biomechanical experiments. However, more precise methods are invasive and impractical, therefore most research is still conducted with optoelectronic commercial systems (which also have some inaccuracies). Furthermore, they compare the accuracy of markerless methods in the HumanEva benchmark \cite{sigal_humaneva:_2010} stating that the precision required for most analysis in sport science is not yet achieved with the current markerless algorithms. While giving a good overview of the field, this survey does not discuss more recent methods (\cite{pavllo_3d_2019}; \cite{cheng20203d}) that produce results with five times better accuracy of their best reviewed methods. It also gathers results from an older dataset and not on the more recent ones such as Human3.6M that contains more images with better resolution.\\

Finally, \cite{chen_monocular_2020} detailed recent monocular 2D and 3D pose estimation techniques that are based on deep learning. They explain in detail the different categories of methods that currently exist and the evaluation protocols and metrics for this task. More than twenty methods are reviewed from 2014 and forward. Our review differs as it focuses on 3D pose estimation, but also considers multi-view settings as well as techniques employing different sensors such as inertial measurement units (IMU).

\subsection{Proposed Approach}
\cite{moeslund_survey_2001} adopts three main criteria to evaluate pose estimation systems\,:
\begin{itemize}
	\item [$\bullet$]Accuracy
	\item [$\bullet$]Speed
	\item [$\bullet$]Robustness
\end{itemize}

The relevance to different fields (surveillance, control or analysis) is then studied. Despite the needs of these domains of application having evolved, this analysis is still relevant. In the present review, our main objective is to guide the choices of developers, engineers and researchers who want to build upon the reviewed algorithms. We compare recently published academic methods with regard to their relevance to different application fields. First, we describe the metrics and benchmarks that are commonly used for method evaluation in section \ref{Evaluation}. Then, we detail and explain the families of architectures used for human pose estimation in section \ref{Architectures}. Next, we present the current state-of-the-art techniques for 3D markerless pose estimation with an emphasis on carefully selected articles in \ref{Methods}. Finally an overall analysis of accuracy, robustness and speed performance indices is available in section \ref{Analysis}. 

\section{Methods Evaluation}
\label{Evaluation}
This section describes how to evaluate markerless pose estimation methods in 3D. To compare these methods, we use evaluations provided across multiple benchmarks datasets. The designers of these datasets often recommend different metrics. We start by describing how these metrics are computed and discuss their relevance in different contexts. For each of these benchmarks the quantity of images, the environments and acquisition modalities are also detailed.

\subsection{Metrics}
\label{Metrics}
Several metrics measure the accuracy of pose estimation algorithms in 3D. Some are processing the average error, other a detection rate with a predefined threshold and finally some uses perceptual or structural criteria. In this subsection, these metrics are described and their strengths and fail cases are discussed in different contexts.\\

\textbf{MPJPE}: Mean Per Joint Position Error. It is one of the most frequently used metrics found in the literature. It is also sometimes referred to as mean reconstruction error or 3D error. MPJPE is the mean of the Euclidean distances between the estimated coordinate and ground truth coordinate over each joint\,:\\

\begin{equation}
MPJPE(x, \hat{x}) = \frac{1}{N}\sum_{i=1}^{N} \left\|m_{\hat{x}}(i) - m_x(i)\right\| \label{eq:1}
\end{equation}\\

N represents the number of processed joints, \( m_{\hat{x}}(i) \) is the function estimating the ith joint coordinates and \( m_x \) is the ground truth position of the joint. Equation \ref{eq:1} is the measurement of MPJPE for one frame and one skeleton. To generalize for video frame the average of each MPJPE's frame of the sequence is calculated.\\

MPJPE is a good baseline metric that can be used to evaluate a wide variety of methods as long as they want to estimate coordinate position and overall skeleton structure. It can also be adapted to evaluate methods that do not estimate the same number of key points as the number of markers used in common datasets and methods estimating relative poses instead of absolute 3D position \cite{sigal_humaneva:_2010}. This is performed by Procrustes alignment (adjustment to the ground truth poses) using a chosen root joint such as the pelvis. It can sometimes be referred as N-MPJPE or P-MPJPE depending if the alignment is by scale only or also by rotation and translation. The main drawbacks of such metrics, identified by \cite{ionescu_human3.6m:_2014}, is their low robustness to outlier errors and the fact that they can be influenced by perceptually irrelevant variations.\\

\textbf{MPJVE}: Mean Per Joint Velocity Error. Introduced by \cite{pavllo_3d_2019}, this metric can be used when a pose sequence is extracted from a video. Here, the absolute position of joints obtained from the MPJPE is insufficient. For this reason, the author use "the MPJPE  of  the  first  derivative  of  the 3D pose sequences" to "measure the smoothness of predictions over time". This technique is useful to compare estimation models that are using temporal data.\\

\textbf{Angular Metrics}: Another approach would be to measure angle errors of joint segments. Ionescu et al. suggest the Mean Per Joint Angle Error (MPJAE) \cite{ionescu_human3.6m:_2014} which is also sometimes referred as Mean Angular Error \cite{agarwal_recovering_2006}:

\begin{equation*}
MPJAE(x, \hat{x}) = \frac{1}{3N}\sum_{i=1}^{3N} |(m_{\hat{x}}(i) - m_x(i) mod \pm 180)| \label{eq:2}
\end{equation*}\\

Here \( m_x \) and \( m_{\hat{x}}(i) \) refer to 3D angles for ground truth and prediction, respectively.

These metrics can be used when the main analysis is performed on angles between two specific limbs rather than the whole body such as for rehabilitation, or sport motion. However, authors of \cite{ionescu_human3.6m:_2014} report that this metric can have little perceptual meaning. It can also be difficult to interpret because angles are calculated locally and joints that are dependent from a faulty predicted one can yield no errors despite a globally misaligned skeleton. As a result, these metrics are less used in recent computer vision publications.\\

\textbf{Thresholds Metrics}: A common approach in 2D Human Pose Estimation and other detection tasks is to define a threshold where a key point is correctly detected. Then, statistics of correctly predicted joints over a set of images can be computed. The percentage of correct parts (PCP) uses half the size of the ground truth segment to determine if a prediction over a limb segment is correct. A 3D version of PCP can also be adapted. A limb is correctly estimated if the following expression is respected:

\begin{equation*}
\frac{\left\| \alpha - \hat{\alpha} \right\| + \left\| \omega - \hat{\omega} \right\|}{2} \leq \theta \left\| \alpha - \omega \right\|
\end{equation*}

\(\alpha\) and \(\omega\) are the two measured coordinates of the extremities of the limb and \(\hat{\alpha}\) and \(\hat{\omega}\) their predictions. \(\theta\) is a chosen parameter to control the accuracy requirement for the threshold (commonly 0.5). This metric was used to evaluate model-based pose estimation using pictorial structure such as \cite{6619308}. The issue with this metric is that a shorter limb will be less likely to be considered detected as the threshold decreases.\\

Another threshold metric used in 2D pose estimation is the Percentage of Correct Keypoints (PCK). This metric does not have the issue of shorter limbs harder to detect as it uses a subject specific threshold for each individual joint instead of limbs. It is calculated with a portion of a fixed limb length (eg: 0.5 times the head bone length, often referred as PCKh@0.5). In this way, the metric is self-adapting to subjects with different proportions, without bias on specific size of the limbs of an individual. \cite{mehta_161109813_2016} propose a 3D version of the PCK used as the main evaluation metric for the MPI-INF-3DHP benchmark. A joint is considered detected with this condition:

\begin{equation*}
\left\| \alpha - \hat{\alpha} \right\| \leq \theta \left\|k - h \right\|
\end{equation*}

\(\alpha\) and \(\hat{\alpha}\) are the target joint and its prediction. \(\theta\) is a parameter controlling the fraction of a reference limb length, \textit{k} and \textit{h} are the coordinates of the extremities of this limb (head, torso...). Another solution is to choose a fixed threshold of 150 mm, which loses the specificity of the subject.\\

Using these threshold-based metrics is justified when comparing methods that could have a good overall accuracy, but produce errors in specific scenario (for singular joints or skeletons). However, this is done at the price of losing sensitivity that could be relevant when analyzing precise local coordinates (at the millimeter scale), as in biomechanical applications.\\

	\begin{table*}[!ht]
		
		\resizebox{\textwidth}{!}{\begin{tabular}{|c|c|c|c|c|c|c|c|c|c|}
				\hline
				Dataset & \#Frame & \#Subject & \#View & Resolution & Frequency & Depth & IMU &  Context and\\
				& \#Video Sequence &  &  & &  &  &  & Main Characteristics\\
				\hline
				Human3.6M: \cite{ionescu_human3.6m:_2014} & 3.6 millions & 11 & 4 & 1000x1000 & 50Hz & Yes & No & Lab environnement  \\
				& 1 376 &  &  & &  &  &  & \\
				\hline
				HumanEva: \cite{sigal_humaneva:_2010}& 80 000 & 4 & 7 & 660x500 & 60Hz & No & No & Lab environnement \\
				& 56 &  &  & &  &  &  & \\
				\hline
				Total Capture: \cite{trumble_total_2017} & 1.9 millions & 5 & 8 & 1920x1080 & 60Hz & No & Yes & Lab environnement\\
				& N/A &  &  & &  &  &  & Inertial Measurement Units\\
				\hline
				&  &  &  &  &  &  &  &"In the wild" \& Lab  \\
				MPI-INF-3DHP: \cite{mehta_161109813_2016}& 1.3 millions & 8 & 14 & N/A & N/A & No & No  &  outdoor/indoor green screens.\\
				& 64 &  &  &  &  &  &  & Markerless ground truthes\\				
				\hline
				&  &  &  &  &  &  &  &Multi-person, "In the wild"\\
				MuPoTS-3D (2018): \cite{mehta_single-shot_2018}& 8 000 & 8 & 1 & 2048x2048 & 30Hz - 60Hz & No & No  &  indoor/outdoor scenes.\\
				& 20 &  &  & 1920x1080 &  &  &  & Markerless ground truthes\\
				\hline
				& &  &  &  &  &  &  &"In the wild" outdoor\\
				3DPW \cite{ferrari_recovering_2018}  & $>50 000$ & 7 & 1 & N/A & 30Hz & No & Yes &single moving camera \& IMUs \\
				& 60& & & & & & &Up to two subjects \\
				\hline
				Carnegie Mellon \cite{cmu_mocap} Mocap &N/A& 109 & 1 & 352x240 & 30Hz & No & No & Indoor environment \\
				& 2 605 & &  &  &  & &  & Various actions and subjects\\
				\hline
				CMU-MMAC: \cite{de_la_torre_guide_2008}& $\approx$450 000 & 25 & 5 & 1024x768 (x3) & 30Hz (x3) & No & Yes & Lab Environnement \\
				& N/A &  &  & 640x480 (x2) & 60Hz (x2) &  &  & Subjects cooking 5 recipes  \\
				\hline
				TNT15: \cite{vonPon2016a}& 13 000 & 4 & 8 & 800x600  & 50Hz & No & Yes & Office environment \\
				& 20 & &  &  &  & &  &No marker-based labeling, only IMU\\
				\hline
				AMASS: \cite{AMASS:2019}& N/A & 346 & variable & variable  & variable & No & No & Unified parametrization of 15 datasets \\
				& ($>$40 hours) & &  &  &  & &  &Mesh body models\\
				\hline
				MoVi: \cite{ghorbani_movi_2020}& N/A & 90 & 4 & 800x600 & 30Hz & No & Yes & Synchronized MoCap \\
				& (17 hours) & &  & 1920x1080 &  & &  & shape, video and IMU data\\
				\hline
		\end{tabular}}		
		\caption{Popular datasets used to compare, train and test human pose estimation models. Video frames, the number of subjects and actions give an indication about the dataset diversity and the number of pose configurations. The number of views from RGB cameras, the resolution and acquisition frequency of cameras assess the quality and quantity of exploitable video information. Inertial Measurement Units (IMU) are sometimes used to refine results from the motion capture or single-image detection. If not specified, the motion capture method is marker-based.}
		\label{tab:datasets}
	\end{table*}

\textbf{Volume \& Surface Based Metrics}\,: Some techniques for human pose estimation need a measurement over surfaces. This type of metric can be found in dense pose estimation Densepose (\cite{guler_densepose_2018}). This task aims at recovering the surface of the whole human body, not just a few joint key points. Geodesic distance-based metrics are often used in this context. An example would be the Geodesic Point Similarity described in \cite{guler_densepose_2018}:

\begin{equation*}
GPS(j) = \frac{1}{|P_j|}\sum_{p \in P_j} \exp \left( \frac{-g(i_p, \hat{i}_p)^2}{2k^2} \right)
\end{equation*}\\

\(P_j\) is a set of points representing the body surface of the jth one person. \(g(i_p, \hat{i}_p)\) is the geodesic distance calculated between the estimated point and the ground truth one. A GPS score of 0.5 indicates that this distance is equal to half a predefined distance adjustable with the k parameter (often setup to be a fraction of a joint segment).\\

3D human shape tracking is another variant of the task that reconstruct and track the human body volume frame by frame in a video. A common approach uses the iterative closest point (ICP) algorithm to fit image data to a model. \cite{huang:hal-01588272} suggest using random forests and nearest-neighbor matching with two volumetric features based on voxels and centroïdal Voronoi tessellation instead of ICP.

Lastly, another popular family of methods is using multi-view data and shape-from-silhouette techniques to create volumetric representations of the human body. This shape can then be useful for joint location prediction. These methods produce probabilistic visual hulls (PVH) (\cite{grauman_bayesian_2003}) in voxel grids. Even though human body shape estimation is not human pose estimation, it is a close task that can be used at different stages of a modular motion capture system. With multi-view datasets it is easy to obtain ground truths PVH that can then be used to evaluate 3D reconstructed volumes (e.g. \cite{trumble_total_2017}). Here Means Squared Error can be calculated from the voxel grid.\\

Each of these metrics can be used in specific variations or edge cases of the task. For "classical" human pose estimation, MPJPE seems more popular as it is simple and no extra parameters intervene in its computation. However, some published articles (\cite{ionescu_human3.6m:_2014}; \cite{mehta_161109813_2016}) claim that threshold metrics are better at identifying errors in specific joints and less prone to penalize perceptually irrelevant errors. Finally, to evaluate methods that process videos and produce 3D pose sequences, the MPJVE is a good alternative to highlight techniques that produce more realistic human motions.\\

Furthermore, the metrics described above only express physical accuracy in multiple ways, with threshold-based ones sometimes introducing perceptual parameters. However, depending on the use case, it might be pertinent to take into consideration more complex perceptual metrics \cite{Marinoiu_Papava_Sminchisescu_2016}, \cite{pictorial_human_space2013} or structural metrics \cite{kocabas_self-supervised_2019}. They can help when purely positional information produce the same error score for two different predicted poses. Significant work has been produced on the way human are perceiving what is a valid and realistic human body configuration. These metrics can be useful in fields that are not concerned about the biological and physical constraints, but more about pose semantic.\\

\subsection{Commonly used Benchmarks}
\label{benchmarks}
Collecting accurate data for human pose estimation is a long and complex process that is driven by progress in acquisition technologies. Moreover, several specific choices are needed concerning the sensor modality, quantity and the acquisition protocol.\\

The complexity of this task explains why it has taken time for the scientific community to create large benchmarks: today many variations exist between monocular versus multi-view, laboratory controlled versus in-the-wild environments (see Table \ref{tab:datasets}) etc. With new commercial solutions (ie: Theia, The Captury) starting to produce results similar to traditional motion capture \cite{kanko_assessment_2020}, some benchmarks are also starting to use markerless labeled ground truths. They as the advantage of easily providing in-the-wild images. However, using this kind of data as ground truth can be questioned as it is itself obtained using methods that are not always available and transparent. Despite this diversity, there are still only a few openly accessible academic benchmarks containing more than millions of images.\\

Datasets have an important role as they are used to validate and test algorithms, but also to train and fine-tune deep learning models. Providing large, high-quality datasets with excellent labels and a wide variety of poses is a major challenge. Three references historically best meet these criteria: HumanEva I and II \cite{sigal_humaneva:_2010}, Human3.6M \cite{ionescu_human3.6m:_2014} and Total Capture \cite{trumble_total_2017}. They contain video sequences with multiple view angles associated with ground-truth joint coordinates. However, these large-scale references are mainly captured in controlled laboratory environments with marker-based systems, hence with a limited variety of backgrounds, poses and subjects. \\

Other benchmarks focus on pose and subject diversity, or on in-the-wild environment acquisition. While interesting to experiment with for human pose estimation or its subtasks, they are not providing the same quantity and variation to conduct large-scale evaluations. However, an noticeable exception is MPI-INF-3DHP \cite{mehta_161109813_2016}, which is more and more used to benchmark algorithms as it proposes a high variety of contexts and subjects (using "green screen" and outdoor acquisition) with a significant amount of data.\\

Marker-based motion capture is the easiest way to obtain something close to ground-truth data. However, old datasets have a low image resolution and some have inaccurate annotation for some subjects (reported by \cite{iskakov_learnable_2019}). Additionally, \cite{colyer_review_2018} noted errors of about 10mm or 10$^{\circ}$ compared to intrusive methods closer to the real human anatomy (ie: intra-cortical bone pins). This is because the key points are reconstructed from groups of markers placed on the subjects' skin or clothing (i.e., soft surfaces).\\

Some datasets propose 3D mesh-based models of the human body (\cite{AMASS:2019}; \cite{ghorbani_movi_2020}). These representations can be used as learning targets for estimation algorithms that care about richer information than skeletal representation of the poses and joint positions. They are also useful for applications that need fully-rigged models such as animation. The AMASS dataset \cite{AMASS:2019} unifies several motion capture datasets by providing the 3D representation of subjects. This is done by computing body poses using the SMPL \cite{loper_smpl_2015} model with their regression method (MoSh++). Furthermore, \cite{ghorbani_movi_2020} provide a new dataset with motion capture (MoCap) and inertial motion unit data, added to AMASS.

\section{Architectures for Human Pose Estimation}
\label{Architectures}
In this section, the main 3D pose estimation families of methods will be described. They can be classified as methods using human body models, learning algorithms or geometric information. In the case of a neural network learning approach, backbone networks are employed and new loss functions are created. Table \ref{tab:taxomethods} is the complete taxonomy of all discussed methods according to these criteria. In the second part of this section, we summarize the most commonly used architectures for 2D and 3D pose estimation.

\begin{table*}[htp]
	\begin{center}
		\resizebox{\textwidth}{!}{\begin{tabular}{|l|c|c|c|c|c|c|c|c|c|c|c|}
				\hline
				\multicolumn{3}{|c|}{} & \multicolumn{3}{|c}{Human Body Models} &	\multicolumn{6}{|c|}{Neural Networks} \\
				\hline
				Method & Proxy Representation & Losses & Kinematic  &Skeleton & Mesh & Backbone & GAN & RNN & TCN & Attention & GCNN \\
				&&&Chains&(e.g. PSM)&(e.g. SMPL)&&Adv.lea.&LSTM&&&\\
				\hline
				\cite{pavlakos_coarse--fine_2017} & 2D Heatmaps & L2 &&&&SHNet&&&&&\\
				\hline
				\cite{mehta_vnect:_2017}&2D heatmaps,& L2&$\bullet$&&&Resnet50&&&&&\\
				&"Location maps"&&&&&&&&&&\\
				\hline
				\cite{zhou_towards_2017}&2D heatmaps&L2,&&&&SHNet&&&&&\\
				&&"geometric loss"&&&&&&&&&\\
				\hline
				\cite{martinez_simple_2017} &2D pose&L2&&&&SHNet (2D)&&&&&\\
				&&&&&&+ MLP&&&&&\\
				\hline
				\cite{sun_integral_2018}&2D/3D heatmaps&any heatmap losses&&&&Resnet models&&&&&\\
				&&&&&&and SHNet tested&&&&&\\
				\hline
				\cite{omran_neural_2018}&part segmentation map&3D and 2D joint loss (L2)&&&$\bullet$&RefineNet \cite{lin_refinenet_2016}&&&&&\\
				&&3D latent parameter loss (L1)&&&&+ Resnet50&&&&&\\
				\hline
				\cite{mehta_single-shot_2018}& "Occlusion Robust Pose Maps" &L2&$\bullet$&&&ResNet50&&&&&\\
				&Part affinity fields&&&&&&&&&&\\				
				\hline
				\cite{kolotouros_learning_2019}& N/A &L2&&&$\bullet$&ResNet50&&&&&\\
				\hline
				\cite{wandt_repnet_2019}&N/A&"Reprojection loss"&$\bullet$&&&SHNet (2D)&$\bullet$&&&&\\
				&&Wasserstein loss, Camera loss&&&&&&&&&\\
				\hline
				\cite{xu_denserac_2019} &pixel-to-surface maps&"render-and-compare loss"&&&$\bullet$&Resnet50&$\bullet$&&&&\\
				&&reconstruction loss (L2), parameter loss (L2)&&&&&&&&&\\
				\hline
				\cite{kocabas_self-supervised_2019}&2D pose&smooth L1&&&&Resnet50&&&&&\\
				\hline
				\cite{mathis_deeplabcut_2018}& N/A &L2&&&&Resnet50&&&&&\\
				&&&&&&Resnet101 tested&&&&&\\
				\hline
				\cite{mehta_xnect_2020}& "3D pose encoding" &smooth L1&$\bullet$&&&"SelecSLS Net"&&&&&\\
				&Part affinity fields&&&&&Fully connected&&&&&\\
				\hline	
				\cite{hossain_exploiting_2018}&2D pose sequence&L2&&&&SHNet (2D)&&$\bullet$&&&\\
				&&derivative loss on joint sets&&&&&&&&&\\
				\hline
				\cite{cai_exploiting_2019}&2D pose, ST-graph&L2, "symmetry loss"&&&&CPN&&&&&$\bullet$\\
				&&derivative loss on joint sets&&&&&&&&&\\
				\hline
				\cite{pavllo_3d_2019}&2D pose&trajectory and pose loss&&&&SHNet, CPN&&&$\bullet$&&\\
				&&bone length L2 loss, 2D projection loss&&&&and Mask-RCNN tested (2D)&&&&&\\
				\hline
				\cite{cheng_occlusion-aware_2019}&2D pose&L2, 2D projection loss&&$\bullet$&&SHNet (2D)&$\bullet$&&$\bullet$&&\\
				\hline
				\cite{cheng20203d}&2D heatmaps&L2, "multi-view loss"&$\bullet$&&&HRNet (2D)&$\bullet$&&$\bullet$&&\\
				&&2D projection loss&&&&&&&&&\\
				\hline
				\cite{liu_attention_2020}&N/A&N/A&&&&SHNet and CPN&&&$\bullet$&$\bullet$&\\
				&&&&&&tested (2D)&&&&&\\
				\hline
				\cite{vedaldi_motion_2020}&2D pose, ST-graph&L2, "motion loss"&&&&CPN and HRNet&&&&&$\bullet$\\
				&&&&&&tested (2D)&&&&&\\				
				\hline
				\cite{qiu_cross_2019}&2D heatmaps&L2&&$\bullet$&&SimpleNet&&&&&\\
				\hline			
				\cite{iskakov_learnable_2019}&2D heatmaps&soft L2, L1 regularized&&&&SimpleNet&&&&&\\
				\hline
				\cite{he_epipolar_2020}&2D pose&L2&&&&SimpleNet&&&&$\bullet$&\\
				\hline
				\cite{vonPon2016a}&multi-view silhouettes&N/A&$\bullet$&&$\bullet$&N/A&&&&&\\
				&IMU orientations&&&&&&&&&&\\
				\hline
				\cite{trumble_total_2017}&PVH, IMU orientations&L2&&&&Classical&&$\bullet$&&&\\
				&then 2D coordinates&&&&&3D CNN&&&&&\\
				\hline
				\cite{ferrari_recovering_2018}&2D pose, IMU orientations&N/A&$\bullet$&&$\bullet$& \cite{cao_realtime_2017} (multi-person)&&&&&\\
				\hline
				\cite{huang_deepfuse_2019}&"multi-channel volume"&L2&&&&SHNet (3D Conv)&&&&&\\
				\hline
				\cite{zhang2020fusing}&2D heatmaps&N/A&&$\bullet$&&SimpleNet&&&&&\\
				\hline
		\end{tabular}}
	\end{center}
	\caption{The taxonomy of reviewed methods. In case of multiple stages we indicate intermediate representations. For learning methods, loss functions and backbone architecture are indicated when they are present (for many two-stage methods these backbones concern only the first stage of 2D detection). Backbones are referred to as \textit{SHNet}: Stacked Hourglass (\cite{newell_stacked_2016}), \textit{CPN}: Cascaded Pyramid (\cite{chen_cascaded_2018}), \textit{HRNet}: High Resolution Network (\cite{sun2019deep}) and \textit{SimpleNet}: Simple Baselines (\cite{xiao2018simple}).}	\label{tab:taxomethods}
\end{table*}

\subsection{3D Pose Estimation Taxonomy}

\subsubsection*{Human Body Models}

Historically, pose estimation algorithms were relying on part-based or  \textbf{skeleton models} of the human body. Each node represented a joint and vertices limb length and orientation. One example of this kind of approach is the Pictorial Structure Model (PSM) introduced by \cite{fischler_representation_1973} and used for pose estimation in \cite{felzenszwalb_pictorial_2005}, \cite{pictorial_human_space2013} and \cite{belagiannis_3d_2014}. Originally, this model is devised as the minimization of an energy function. The cost function combines an error term for joint location error and a penalty for segment length deformations (i.e. not corresponding to limb size).\\

Several methods adopt \textbf{kinematic} based human skeleton, where each linked joint pair is represented as a vector. With this technique, angular and length constraints can be applied to detect poses. Kinematic Chain Space proposed by \cite{wandt_repnet_2019} is an example of such techniques.\\

\textbf{Mesh models} consists of complete reconstruction of the human body surface. These models offer a richer information than skeleton-based models and can be used to infer the spatial representation of a subject in a virtual scene or to render captured pose into fully-rigged meshes for animation. SMPL \cite{loper_smpl_2015} is the most frequently used human mesh model for pose estimation. The mesh is learned from numerous 3D body scans and can be adapted to a set of pose and shape parameters produced by pose estimation algorithms.

\subsubsection{Geometric Information}
When several cameras are available, multi-view geometry is frequently used for 3D pose estimation. One way to infer joint coordinates in three dimensions is to use \textbf{triangulation} with their 2D image coordinates in each view. Depending on the calibration and availability of camera extrinsic and intrinsic parameters, different reconstruction schemes are possible.\\

Another approach consists of \textbf{fusing features} from different views along epipolar lines before inferring poses. The epipolar line is the image in one camera of a ray passing through the optical center of the other camera and a point in the scene. Considering a point in the first view, its corresponding point in the second view is guaranteed to lie on the epipolar line in the other image. Using this prior information about the different views, the 2D pose can be refined or multi-view features can be merged before the 3D pose itself is estimated.\\

Finally, other methods use shape-from-silhouette reconstruction to obtain the whole body shape before joint detection. These techniques first segment human shapes from each view and then reconstruct their volume (an example of these methods are probabilistic visual hulls from \cite{grauman_bayesian_2003}). These volumes can then be used as intermediate features.

\subsubsection*{Learning Approaches}
Since 2014 with \cite{toshev_deeppose:_2014} and \cite{tompson_joint_2014} convolutional neural networks were extensively used for human pose estimation. However, new architectures and training reformulations of their different building blocks are presented each year. These variants sometimes contribute to the improvement of the state-of-the-art for 3D pose estimation. Here, we review the most commonly used families of networks for this task.\\

The first thing to consider is the design choices regarding the convolution operations. For 3D pose estimation, many variants are employed according to the data representation that is used or the hypothesis of the authors. The choices range from \textbf{2D convolutions} on image data to convolutions on spatial-temporal graph representing a subject motion. Most monocular methods commonly use classic convolutions. For video sequences, multi-view or multimodal setups, richer information have inspired different techniques. Volumetric intermediate features computed from multi-view can be fed to \textbf{3D convolution} networks to refine or estimate the pose. \textbf{Temporal convolutions} can be used to reason on past and future frames and help better characterization of the pose.\\

Sometimes, the location of the pose and the trajectory in time are encoded by spatial-temporal graph that can be computed with \textbf{Graph Neural Network} (GNN). A spectral convolution can then be applied in the Fourrier domain using the eigenvalues of the Laplacian graph \cite{kipf_semi-supervised_2017}.\\

\textbf{Recurrent Neural Networks} (RNN) are another way to process pose sequences. More specifically \textbf{Long Short-Term Memory} (LSTM) architectures have been used with success on 2D joint sequences \cite{hossain_exploiting_2018} and other modalities \cite{trumble_total_2017}. This technique showed success in text translation and other tasks to process sequences with long-term dependencies. From one sequence, LSTM can output another one keeping information about previous inputs passed successively (sequence-to-sequence model). LSTMs are distinguished from RNNs by the state of their cells that is updated by different linear operations called "gates." These operations select and update information that are useful to remember (this website from \href{https://colah.github.io/posts/2015-08-Understanding-LSTMs/}{Christopher} \cite{noauthor_understanding_nodate} details LSTM functioning).\\

\textbf{Attention mechanism} aims to focus networks on the most important information for pose estimation in the input data or intermediate features. \cite{cai_exploiting_2019} use attention to select the frames that contribute the most to the estimation, whereas \cite{he_epipolar_2020} describe an "Epipolar Transformer" module taking advantage of multi-view to focus on learning across epipolar lines. Features in the paired image along the epipolar lines corresponding to the joint points are used.\\

Finally, another interesting line of research concerns generative adversarial networks and \textbf{adversarial learning} (\cite{goodfellow_generative_2014}). It has mainly been used for 3D pose estimation in two ways: unsupervised mapping of 2D to 3D poses \cite{kudo_unsupervised_2018} and more frequently as a pose validation module (\cite{cheng20203d}; \cite{wandt_repnet_2019}; \cite{kocabas_vibe:_2019}). In this case, a discriminator network improves pose consistency by being trained to recognize generated poses from the ones directly extracted from ground-truths. An adversarial loss component is then propagated to the generator network which estimates 3D human poses from visual information or 2D poses. Thus, poses that are inconsistent with known configurations are penalized.

\subsection{Backbone Architectures}

\begin{figure*}[ht]
	\begin{center}
		\includegraphics[height=0.4\textheight]{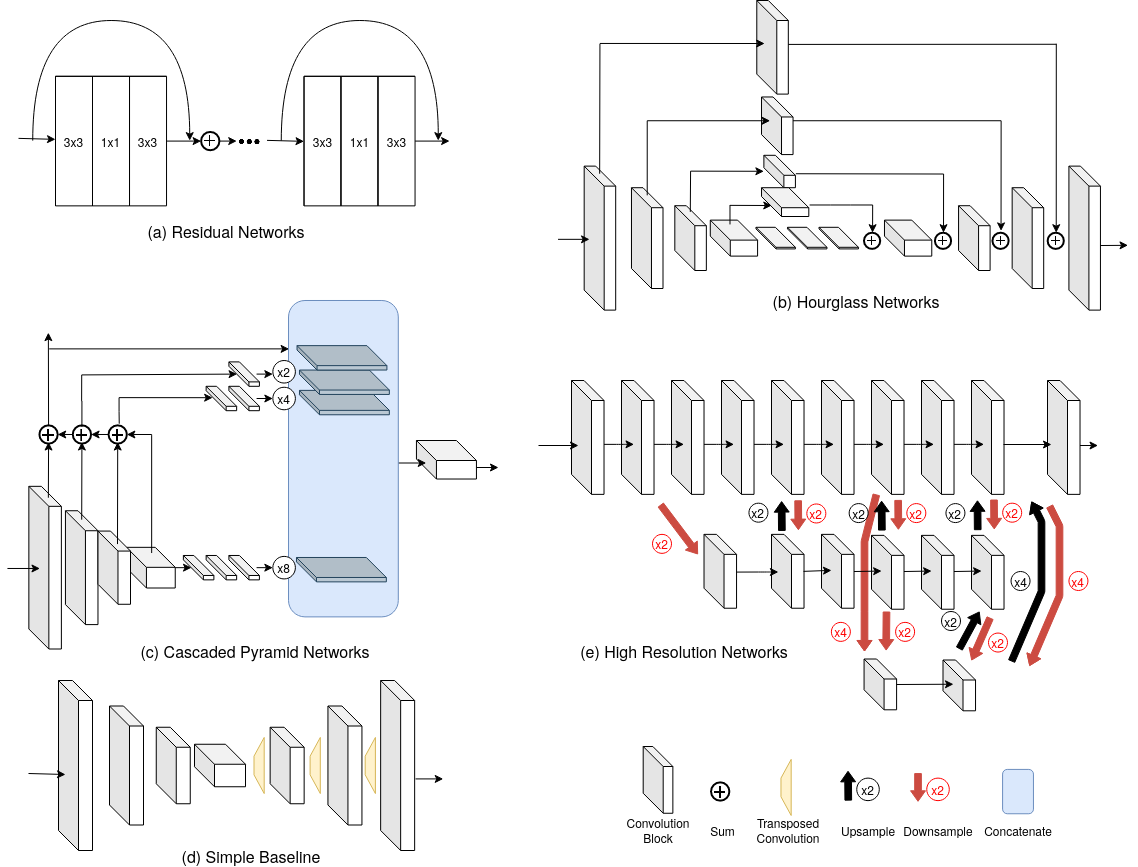}
	\end{center}
	\caption{ Backbone architectures used for 2D pose estimation. (a) Residual blocks are the main characteristic of ResNet variants (Resnet101, Resnet50, Resnet152) \cite{he_deep_2015}. They are also present in more "human pose estimation" specialized models: (b) Stacked Hourglass networks \cite{newell_stacked_2016} and (c) cascaded pyramid networks \cite{chen_cascaded_2018}. The simple baseline (d) network proposed by \cite{xiao2018simple} uses transposed convolutions to recover the higher input resolution. (e) Higher resolution network (HRNet) \cite{sun2019deep} processes high and low resolutions features in parallel within sub-networks that share information.}
	\label{fig:Archi2D}
\end{figure*}

\label{2DNetworkArchis}
Before reviewing state-of-the-art algorithms it is important to consider the backbone architectures that are commonly used for 2D or 3D pose estimation. They are extensively used in "top-down" approaches before any computation reducing the problem to a mapping of 2D to 3D coordinates. They reason from low-level joint coordinates to infer high-level information about the human skeleton (see Fig. \ref{fig:3DMarkerlessSystems}). In opposition, "bottom-up" techniques reason on human body models and extract features from the images after fitting them to the data. These backbone architectures are also employed directly for 3D pose estimation. For example, \cite{huang_deepfuse_2019} use Hourglass Networks including 3D convolution on volumetric representations. Space is missing to explain the backbones architectures methods in details, but the reader can refer to \cite{xiao2018simple}, \cite{newell_stacked_2016} or \cite{chen_cascaded_2018}. In the present paper, we limit ourselves to the general principles of commonly used backbone architectures and provide an overview of their structure and performance.\\

Two commonly used architectures are hourglass networks \cite{newell_stacked_2016} and cascaded pyramid networks \cite{chen_cascaded_2018} (Fig. \ref{fig:Archi2D}, (b) and (c)). They both compute image features at different resolution level with the key idea to encompass global and local discriminant features. Hourglass networks are composed of stacked hourglass modules\,: the first part of the module is reducing the resolution as it passes through convolutional layers. The second part up-samples the features while summing them with corresponding ones of the same dimension from the previous stage. Intermediate supervision is conducted at the end of each module. Cascaded pyramid networks are a two-step architecture predicting poses from a feature pyramid network. It then refine the results for hard-to-predict points. The pyramid network fuses the features at different resolutions to produce joint position heat maps. In the second stage, the refinement process is using intermediate features from the pyramid at different levels. They are up-sampled and summed before going through a final convolutional block. This process is done only for "difficult to predict key points" (chosen with the loss from the pyramid network). These two architectures are using residual connection modules \cite{he_deep_2015} as "building blocks". Based on skip connection between convolutional layers, this technique is widely used in many computer vision tasks for feature extraction. Some 2D pose estimation systems are directly using one of the original variants of this network (ie: Resnet50, Resnet101 etc).\\

High resolution networks \cite{sun2019deep} are considered as another architecture exploiting different resolutions of the same image. Here, it processes resolutions in parallel with convolution layers sharing weights through "exchange blocks."\\

The efficient 2D multi-person detector from \cite{cao_realtime_2017} is sometime used. It uses "part affinity fields" which is "a non-parametric representation of relationship between body parts". From these features and joint localization confidence maps, human poses are predicted with correct associations to multiple subjects. Another asset of this method is that it runs in real-time.\\

Finally, \cite{xiao2018simple} propose a baseline method also using Resnet as its base. This technique is using deconvolution layers to produce heat maps from deep image features, without a specific procedure for difficult-to-predict points. Despite its simplicity, this model achieves competitive results at an efficient computational cost. It is therefore used for comparison evaluations but also sometimes as a backbone 2D detector.
\section{Methods Review}
\label{Methods}

This section describes and compares the top performing methods for vision based 3D markerless pose estimation (see Tables \ref{tab:methods-monocular}, \ref{tab:methods-temporal} and \ref{tab:methods-multimodal}). The selection process was as the follows\,:\\

\begin{itemize}
	\item Top methods from the state-of-the-art on each most popular benchmark
	\item Most cited methods in computer vision and fields of application for 3D human pose estimation (biomechanics, robotics, sport sciences, human-machine interaction etc.)
	\item Recent methods reporting interesting results or original approaches that can further advance research
\end{itemize}

The first observation regarding the evolution of accuracy in the state-of-the-art is that it is rapidly improving. This task is getting attention and, every year new, records are reached. The average error dropped from about 100mm to less than 20mm within 10 years. It is yet to be determined whether accuracy is still going to increase in the future. However, given the diversity of approaches and modalities, it can be argued that, to date, there is no consensus on the best method to use.\\

In this review, we focus on the best performing markerless techniques currently available, regardless of the sensors or algorithms they use. Methods using monocular images and video sequences are first presented. Second, we present methods using multiple views. Finally, we include methods using other modalities as a pre-processing step, or during the prediction, specifically inertial measurement units. An overall analysis of the performance indices of the different families of methods will be conducted in the following sections.\\

\subsection{Monocular Images}

\begin{table*}[htp]
	\begin{center}
		\begin{tabular}{|l|c|c|c|}
			\hline
			Method & Human3.6M & MPI-INF-3DHP &  HumanEva\\
			\hline\hline
			\cite{pavlakos_coarse--fine_2017} & 71.90 & - & \textbf{24.3}\\
			\cite{mehta_vnect:_2017}&80.5*&76.6&-\\
			\cite{zhou_towards_2017}&64.9&69.2&-\\			
			\cite{martinez_simple_2017} & 62.9/47.7* &-&24.6\\
			\cite{sun_integral_2018}&64.1/49.6+/40.6*+&-&-\\
			\cite{omran_neural_2018}&59.9*& - &64.0*\\
			\cite{mehta_single-shot_2018} & 69.9 & 74.1 & - \\
			\cite{kolotouros_learning_2019}&\textbf{41.1}&76.4&-\\
			\cite{wandt_repnet_2019}&50.9 / 38.2*&82.5&-\\
			\cite{xu_denserac_2019} & 82.4 / 53.9* /48.0*+ & 73.1 / 76.9+ / 89.0*+& - \\
			\cite{kocabas_self-supervised_2019}&51.83+/45.04*+&77.5&-\\
			\cite{mathis_deeplabcut_2018} (DeepLabCut)&-&-&-\\
			\cite{mehta_xnect_2020} & 63.6 & \textbf{82.8} & - \\
			\hline
		\end{tabular}
	\end{center}
	\caption{Accuracy comparison from several state-of-the-art monocular methods. Human3.6M and HumanEva results reported in absolute MPJPE (lower is better); Results from MPI-INF-3DHP are reported in 3DPCK (higher is better). Techniques with the annotation with + are using extra-training data to obtain the result; the others use the benchmark's recommended protocols. The * annotation indicates results published with procrustes alignment to ground truth poses before evaluation.}
	\label{tab:methods-monocular}
\end{table*}

\cite{pavlakos_coarse--fine_2017} present one of the first methods to propose a one-stage end-to-end convolutional neural network to predict 3D human pose from a single RGB image. They do so by focusing on the 3D nature of the task. Their architecture uses stacked hourglass modules \cite{newell_stacked_2016} (see 2D architectures \ref{2DNetworkArchis}) that outputs volumes of voxel probability for each joint in a 3D discretized space around the target. They also propose a new intermediate supervision method inspired from success in 2D human pose estimation. This original method does not use a 2D joint estimation step. Instead, they employ a coarse-to-fine approach that leverage 2D heatmaps as ground truths for intermediate supervision, and then fuse them with image features as an output of the next modules. Further down, the network reconstructs 3D voxel maps. The supervision is also done using 3D Gaussian around the given 3D coordinates ground truths. This deep network provides accurate results (72mm) for a monocular method using purely 3D data representation. However, more recent methods showing better results use two-stages top-down architectures including 2D prediction as a first step for 3D detection. This suggests that image features are not rich enough for direct 3D inferences.\\

VNect from \cite{mehta_vnect:_2017} is a framework for 3D root-relative human pose estimation in real-time. It consists of a similar to Resnet50 architecture that generates 2D heat maps for each joint as well as newly introduced "location maps." These location maps predict the relative X, Y, and Z positions of an articulation relative to their root joint. For each joint, the location is processed from the peek of the 2D heat maps and the root-relative coordinate is read in the location maps. Then, a kinematic model of the human skeleton is fitted to the predicted poses, to improve temporal consistency and reduce jitter. The strengths of this method are that it works in real-time and can be used in different outdoor contexts. However, the authors list several limitations to their approach, mainly that depth estimation errors might lead to erroneous 3D predictions. The method is also only capable of relative pose estimation and therefore requires accurate detection of a root joint (pelvis).\\

\cite{mehta_single-shot_2018} following Vnect, this technique also uses location maps but modify them to become "Occlusion-Robust Pose-Maps" (ORPM) that infers 3D pose from multiple subjects. These ORPMs are similar to location maps but also contain structural information about the pose. For each joint, the location in the 3D maps is stored at the position of the joint but also along a predefined set of joints (dividing the body into two arms and legs) and root joints (pelvis and neck). They also employ part affinity fields \cite{cao_realtime_2017}, which are often used in 2D multi-person pose estimation. They represent "2D vector fields pointing from a joint [...] to its parents.” Thus, starting from valid root joints, all joints can be inferred following the kinematic chain of the human body. The outstanding feature of this method is that it perform multi-person detection and in a single-shot manner without intermediate use of off-the-shelf 2D pose estimators or person detector. It also introduces two new dataset for training and evaluation: MuCo-3DHP (large-scale multi-person with occlusion dataset, composed from MPI-INF-3D images) and MuPoTS-3D (In-the-wild multi-person dataset, filmed in various environments with markerless ground truths).

\cite{mehta_xnect_2020} More recently, in continuation of the previous two methods, the Xnect framework was presented. It is athree-stages method that combines a convolutional network for feature extraction, a fully connected network for pose estimation and a fit of the previous results to a kinematic model to refine pose consistency. The first stageextracts 2D and subject association through part affinity fields (similar to \cite{cao_realtime_2017}) as well as 3D pose encoding in the same way as the previous methods. The key difference from \cite{mehta_single-shot_2018} is that each joint only encodes information about its position relative to the parent joint and the position of its children. For this stage, a new backbone module is presented for network architectures to reduce computational costs: SelecSLS. It consists of successive 3 by 3 and 1 by 1 convolutions with inter- and intra-module skip connections. It achieves similar accuracy to Resnet50 while being 1.4 faster at inference. The second stage consists of 3D pose detection, for each subject in parallel, through fully connected networks trained on the MuCo-3DHP dataset incorporating examples with severe occlusions. Finally, kinematic skeleton fitting is performed using the minimization of an energy term consisting of position (through inverse kinematics for 3D features and 2D re-projection), orientation and temporal consistency. Clearly, this is a complex framework that involves many steps, as well as re-tracking of each subject before the last stage. However, the framework is robust to occlusions and adapted to multi-person. It is also computationally efficient since it can run in real-time at 30 fps on generic hardware configurations. However, it appears that the last stage of the framework decreases the overall accuracy while performing better on certain joints and producing a smoother orientation estimate.

\cite{zhou_towards_2017} published a two-stage method using the hourglass network architecture of \cite{newell_stacked_2016} for 2D heat map generation, then regress depth for each joint. In addition, they apply a weakly-supervised process to exploit images that have only been labeled with 2D ground truths. The depth prediction is realized from features at different resolutions, which are extracted from several levels of the Hourglass network. The weakly-supervised training is applied using 3D and 2D labeled data. Euclidean loss is applied to predictions on images with 3D ground truths, whereas a geometric loss is used when only 2D labels are available. This lossadds constraints from the average limb length ratios among predefined bone-groups. The main contribution of this work is the weakly-supervised technique: the authors evaluated whether the contribution of 2D data improved results on the 2D portion of the framework or on the whole 3D estimation task. They state that at PCKh@0.5 (standard metric for 2D pose estimation see \ref{Metrics}), the results are similar for 2D human pose estimation when 2D data is not used, but depth estimation is greatly improved. However, an analysis with a threshold smaller than 0.5 might be interesting,  as a small increase in 2D accuracy is still observed. Nevertheless, their work confirms that using 2D detectors as part of the 3D detection task is possible and yields accurate results ($\simeq$65mm MPJPE).\\

\cite{martinez_simple_2017} presented simple baseline for 3D human pose estimation. This method differs from others in that it does not use image data or intermediate feature maps (ie. joint location heat maps) nor does it use optimization steps with model fitting. Instead, it infers 3D coordinates from 2D coordinates obtained with a state-of-the-art 2D human pose estimation architecture. Despite this simple design, it produces accurate results comparable to and sometimes better than some more complex contemporary techniques. The design choices for the model are the following: a simple 2-layer CNN with batch normalization, ReLU activation, and dropout. It takes as input the 2D predictions of the Hourglass Network \cite{newell_stacked_2016}. The results obtained on Human3.6M reach 62mm average error on single images. This method is also one of the first to propose the direct lifting of 2D coordinates from efficient 2D detectors to 3D. The high accuracy obtained with such a simple approach without image data as an input lead the authors to hypothesize that the visual features used in contemporary methods were either not useful to 3D human pose estimation or still under-exploited. The latter hypothesis tends to be confirmed with new methods achieving increasing accuracy with clever exploitation of temporal features (\cite{hossain_exploiting_2018}; \cite{pavllo_3d_2019}; \cite{cheng_occlusion-aware_2019}; \cite{cheng20203d}) or combination of direct regression and model fitting.Because it is simple, fast and is driven by the increasing performances of 2D detectors, this method and the two-stage technique inspired many recent studies.\\

\cite{sun_integral_2018} introduced "integral regression" to extract 3D coordinates from 2D confidence heat maps. It is a function similar to the soft-argmax function commonly used in classification to normalize outputs. Here, it is used to switch from 2D pixel maps to differentiable coordinates, allowing direct regression within an end-to-end network. Their article describes extensive ablation studies on different training architectures losses and backbone (hourglass and residual networks \ref{2DNetworkArchis}) and presents good results for both 2D and 3D human pose estimation. The authors also adopt a training strategy \cite{sun_compositional_2017} allowing the usage of 2D labeled data with separate supervision for the xy coordinates and for the depth, achieving even higher accuracy on Human3.6M. Many approaches are now using 2D heat maps with the soft-argmax regression.

\cite{omran_neural_2018} present the Neural Body Fitting approach that combines part segmentation with a convolutional neural network and a parameterized human body model (SMPL \cite{loper_smpl_2015}). The first stage predicts a part segmentation map using the RefineNet \cite{lin_refinenet_2016} CNN architecture. In the second stage, these part masks are directly fed to a ResNet-50 network that estimates the pose and shape parameter of the SMPL model. The whole process is fully differentiable and can also be trained with 2D data in a weakly supervised manner. The authors demonstrate this by re-projecting the joint coordinate on 2D images and find that, with only 20\% of the training data with 3D ground truths, the same accuracy is achieved as with complete annotations. This method is one of the first that describes the training of CNNs followed with the fitting to a parameterized human body model in a single integrated framework. The other characteristic that differentiates this method from others is that it uses the intermediate feature of the part segmentation map. According to the author analysis, using a 24-part segmentation as an input led to significantly better results than direct image data or joint coordinates. This result is particularly interesting and should be considered when designing architectures that want to reason on the 3D structure of the human body.

Similarly, \cite{kolotouros_learning_2019} designed a system with joint usage of a CNN for direct key point regression and an iterative model optimization technique using a human volumetric model (SMPL \cite{loper_smpl_2015}). The neural network produces good initialization for the iterative fitting of the human model. Then, once the shape model position is refined, it is used to calculate a loss from the initial prediction, which increases the accuracy of the network. The originality of the method lies in using the best of the two techniques. Direct regression allows fast initialization without a priori knowledge directly on the image data; iterative optimization from a human model produces a better shape for the image fitting. The two parts of the system complement and improve each other during each training cycle. The results obtained using this method on Human3.6M are the most accurate of monocular non temporal methods (on a single isolated RGB image). The average error is 41.1mm, which is close to the accuracy of temporal methods. These results show that with simple input data and a suitable method, high accuracy can be achieved. The question is: Could this direct regression/model fitting method using only visual features be suplemented with temporal data to reach a higher accuracy score?

\cite{wandt_repnet_2019} uses the kinematic chain space to represent human 3D pose within their discriminator (or "critic") network. To project pose coordinates in the kinematic chain space, limbs (i.e., edges between detected joints) are described as directional vectors. It is then possible to map this representation back to point coordinates. The kinematic chain representation contains information about limb length, angles and body symmetry, while being easy to compute. Later, \cite{cheng20203d} used a temporal version of the kinematic chain space that is reporting the angle and length modification across frames in a video.\\

\cite{xu_denserac_2019} proposed the DenseRaC framework that converts pixel-to-surface maps of the human body (IUV) into parameters for statistical human body model. Similarly to \cite{omran_neural_2018}, this framework uses intermediate features but, instead of part segmentation, DenseRaC uses pixel-to-3D surface maps. In the first stage of the pipeline, these maps are computed using the Densepose-RCNN architecture (a network that is trained on Densepose-COCO, a dataset of manually annotated human body surface \cite{guler_densepose_2018}). To improve training, the authors present a large-scale dataset of synthetic human poses that can also easily produce IUVs. In the second stage, the IUVs are fed to a regression network that estimates the human body shape and pose parameters ( similar to \cite{loper_smpl_2015}) as well as the camera parameters. Once reconstructed, the 3D mesh is re-projected using a differentiable renderer and rasterized in an IUV similar to those produced in the first step. Then, the computation of an adversarial loss with a discriminator network helps to eliminate impossible configurations.

\cite{kocabas_self-supervised_2019} takes advantage of the multi-view setting that each major motion capture dataset provides. the authors propose to infer geometry from matching 2D detection in each view and then deduce the 3D coordinates. This technique renders self-supervision possible with completely unlabeled data. Unlike many multi-view techniques (\cite{iskakov_learnable_2019}; \cite{he_epipolar_2020}) this method does not use camera parameters. It's important to note that the training part of the network uses a multi-view setting, but during prediction it becomes a monocular method. The architecture is composed of two networks using ResNet50 and a deconvolution layer as their backbone (see 2D architectures \ref{2DNetworkArchis}). Both produce spatial heat maps for each joint in each view. The difference is that, in one case, they are converted in 2D coordinates and in 3D in the other (both with a soft argmax). The 3D network is the one that will be used for inferences, and that has learnable weights. The 2D network is frozen and will be used for the self-supervision.Although the cameras are not calibrated, the authors advise using detected the joint key points in each view to obtain the camera parameters (using RANSAC and SVD). Then, triangulation can be performed to obtain the 3D coordinates, which are then used to supervise the 3D network with a smooth absolute loss. The accuracy score of the method is the current best when few or no labeled data are available (70mm average MPJPE on H3.6M and 64.7 3DPCK on MPI-INF-3DHP). This learning scheme is promising and allows for high accuracy by requiring only raw data (discounting pre-trained 2D detector), provided that multi-view images are accessible.\\

Finally, it is important to consider the methods being used in several research fields. The main one, DeepLabCut \cite{mathis_deeplabcut_2018}, \cite{mathis_deep_2019}, \cite{nath_using_2018} is a generic markerless keypoint-tracking framework developed with the goal to obtain accuracy results similar to human annotation. Its original application target was the video tracking of predefined keypoints for different species. DeepLabCut achieves good results with little training data, which has made it popular and it is now cited in numerous articles in neuroscience and human movement research. For human gait analysis \cite{10.1145/3341105.3373963}{FIKER2020108775}, its accuracy is similar to marker-based motion capture. Although DeepLabCut is not in the strict sense a HPE model, it can be used as one. In fact, it is based on an HPE architecture: It consists of the first layers of the DeeperCut network {DBLP:journals/corr/InsafutdinovPAA16}, which is a multi-person 2D pose estimation model used here for feature extraction, followed by a deconvolution layer. This model and its inspiration do not directly fall within the scope of 3D human pose estimation. Yet, DeepLabCut has been used in a multi-view calibrated cameras context with simple triangulation (\cite{nath_using_2018}; \cite{sheshadri_3d_2020}). As this method seems popular in different fields for its flexibility and performance in a wide variety of contexts, it is important to note that more recent results on specific 2D keypoint detection have outperformed DeeperCut \cite{newell_stacked_2016}. The main appeal for this method is the definition of personalized labels and its powerful generalization capability.\\

\begin{figure*}[ht]
	\begin{center}
		\includegraphics[height=0.5\textheight]{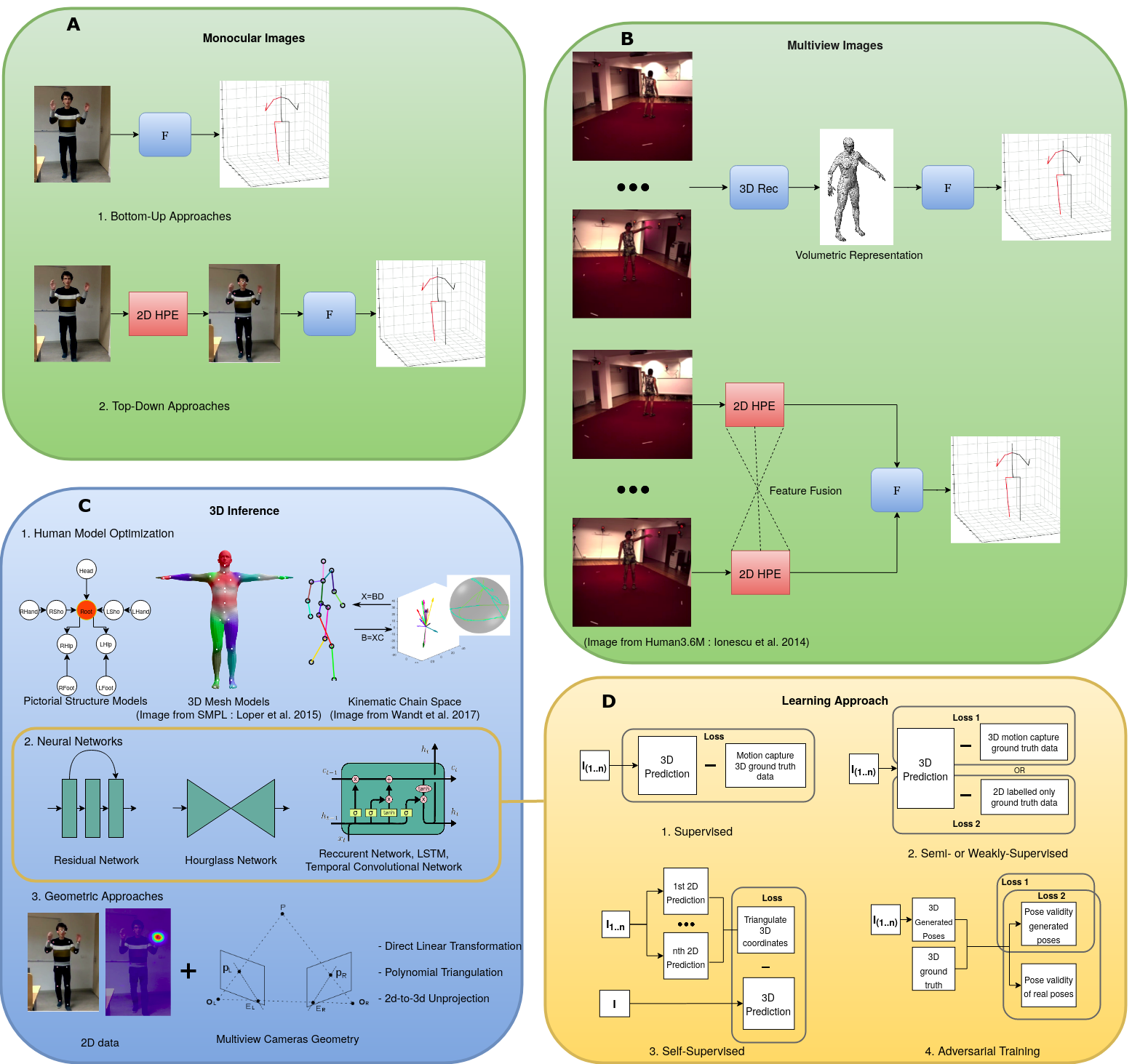}
	\end{center}
	\caption{Overview of the different levels of 3D markerless human pose estimation. \textbf{A}: Monocular approaches, commonly used 2D pose estimation backbone architectures are described in \ref{2DNetworkArchis}. \textbf{B}: 3D feature exploitation and multi-view 2D detection as an input for 3D detectors. \textbf{C}: The different families of 3D pose estimation. \textbf{D}: examples of learning approaches applied to human pose estimation.}
	\label{fig:3DMarkerlessSystems}
\end{figure*}

\subsection{Temporal and Monocular Images}
\begin{table*}[htp]
	\begin{center}
		\begin{tabular}{|l|c|c|c|}
				\hline
				Method & Human3.6M & MPI-INF-3DHP &  HumanEva \\
				\hline\hline
				\cite{hossain_exploiting_2018}  & 58.5 & - & 22.0\\
				\cite{cai_exploiting_2019}&48.8 / 39.0*& - & - \\
				\cite{pavllo_3d_2019}&46.8 / 36.5*&-&23.1/15.8*\\
				\cite{cheng_occlusion-aware_2019} & 42.9/32.8* & - & 14.3*\\
				\cite{cheng20203d} & \textbf{40.1} & \textbf{84.1} & \textbf{13.5*}\\
				\cite{liu_attention_2020}&45.1 / 35.6*& - & 15.4*\\
				\cite{vedaldi_motion_2020}&42.6 / 32.7*&86.9(2d GT)& -\\
				\hline
		\end{tabular}
	\end{center}
	\caption{Accuracy comparison from several state-of-the-art monocular temporal methods. Human3.6M and HumanEva results reported in absolute MPJPE (lower is better); Results from MPI-INF-3DHP are reported in 3DPCK (higher is better). Techniques with the annotation with + are using extra-training data to obtain the result; the others use the benchmark's recommended protocols. The * annotation indicates results published with procrustes alignment to ground truth poses before evaluation. }
	\label{tab:methods-temporal}
\end{table*}

\cite{hossain_exploiting_2018} used the sequence-to-sequence architecture that was initially applied in machine translation tasks for human pose estimation. The main idea is to use long-term temporal context to predict a new sequence, as for the translation of text into another language. Here, 2D coordinate sequences (predicted with the Hourglass Network architecture) are used to predict the 3D ones using Long Short-Term Memory units (LSTM). The first layer of the network encodes the sequences into a hidden space and the last layers decodes them into a 3D dimensional sequence using residual connections. Additionally, important choices are made for the training. First, the first derivative of the 3D coordinates is included in the loss to mitigate the effect of errors during the 2D prediction stage by enforcing temporal consistency. Second, the training combines layer normalization, recurrent dropout and finally the weighting of joint sets, to force the network to better predict challenging parts. To date, the method is one of the most accurate among methods using temporal data. Notably, an evaluation by the authors shows that the optimal sequence length for their model is 5 frames. Beyond this point, the accuracy slowly decreases, suggesting that either the long-term temporal context does not provide good information or the model does not properly exploit past image data.One problem with this architecture is the fixed length of the temporal data input, which was later solved using temporal convolutional networks (TCNs) \cite{pavllo_3d_2019} \cite{cheng20203d}. This paper further details and analyzes other useful techniques applied to temporal pose data, such as the important contribution of residual connections.\\

\cite{cai_exploiting_2019} also uses temporal dependency within the pose sequence that composes the subject motion. However, in addition to time constraints, joint position constraints are also enforced using a graph structure. Edges are present to model spatial continuity and symmetry, but also connections with the same joint in future and past frames of the video. These graphs are computed from 2D skeletons that are then passed to a graph convolutional network. Depending on the type of neighboring (various spatial constraints and temporal connections), the authors propose a different type of convolution. Then, a local-to-global architecture is used similarly to \cite{newell_stacked_2016} to obtain 3D predictions. The graph structure is pooled and upsampled using lower level features from the previous layers in a hierarchical manner. Then, the method produces a 3D pose from a pose refinement module consisting of a fully connected layer. The authors employ a loss strategy combining different terms to enforce joint locations, limb symmetry and temporal smoothness. The results obtained on different benchmarks show a better accuracy than methods based purely on temporal sequences. However, the method is likely not suitable for sequences shorter than 7 frames on most actions. It is also important to note that the effect of the chosen 2D detector is not tested.

As mentioned previously \cite{pavllo_3d_2019}, the technique consists of a monocular method that leverages a fully convolutional network to process temporal data. The main advantage is that, compared to RNN and LSTM, they are more computationally efficient and do not need a fixed input size.In this paper, the authors present an architecture based on time-dilated convolutions that also apply semi-supervision to add non-labeled data to the training. The dilated convolutions capture the long-term dependencies, but also need less training parameters and have a better computation speed than sequence-to-sequence models \cite{hossain_exploiting_2018}. The authors present a comparison showing that about the same number of parameters and FLOPs are needed to yield better accuracy. The semi-supervised process they used is called "back-projection": an encoder-decoder architecture encodes a 2D joint prediction into 3D pose and decodes by projecting it back in 2D. The error is then computed between the original and back-projected 2D predictions.This method has a 10\% increase in accuracy on two of the main benchmarks, leading to an average error of less than 50mm for monocular methods.The methods that follow this paper, based on video sequences, still use the Temporal Convolutional Networks (TCN) architecture (\cite{cheng_occlusion-aware_2019}, \cite{cheng20203d}, \cite{liu_attention_2020}). Another interesting result is the impact of the 2D detector used. The tests conducted show that the CPN and the Mask-RCNN 2D detectors give better results in the end with their model. The authors think that "it's due to the higher heatmap resolution,  stronger feature combination" from these detectors.\\

\cite{cheng_occlusion-aware_2019} and \cite{cheng20203d} also use temporal convolutional networks with a specific occlusion training to improve prediction on challenging images.

In \cite{cheng_occlusion-aware_2019}, they describe an occlusion-aware network where a "cylinder man model" is produces occlusion labels. Their architecture is composed of an end-to-end trainable human detector, a 2D pose estimator, a 3D pose estimator and finally a pose discriminator. The two pose estimation algorithms are respectively a 2D and a 3D temporal convolutional networks. They also use 2D-only data during training using a re-projection loss such as in \cite{pavllo_3d_2019}. The occlusion model intervenes at this stage where self-occluded points (computed with the cylinder model) are not taken into account as they are deemed unreliable.\\

Building on the previous work, \cite{cheng20203d} describe an end-to-end trainable model with several modules that reconstruct 3D joints from monocular videos. 2D confidence heat maps are estimated and used as a feature for 3D prediction. They use a multi-scale convolutional network (HRNet) that fuses spatial features \cite{sun2019deep}. Thus, the main characteristics of their method are the following\,:\\

\begin{itemize}
	
	\item Learning of a multi-scale embedding obtained from those heat maps. 3D poses are then predicted with the embedding using a temporal convolutional network (TCN) \cite{pavllo_3d_2019}.
	
	\item Validation of the pose sequences with a discriminant model based on Spatio-Temporal Kinematic Chains (which enforces limbs angular and length  constraints).
	
	\item  Data augmentation using synthetic occlusion at different levels during TCN training.
	
\end{itemize}

Semi-supervised learning is used with the goal of including strictly 2D labeled data during the training process as in \cite{pavllo_3d_2019}. Finally, and only for the training stage, multi-view from Human3.6M dataset is used to enforce a good skeleton orientation prediction. To do so, the authors use a loss function comparing each inference from two pairs of different views at the same time in the video (after applying a camera rotation known from calibration data). Among the methods reviewed here, this method achieves the overall best results on HumanEVA and MPI-INF-3DHP. It also has the best results for a monocular method on Human3.6M (40mm MPJPE). However, it is a complex method with many submodules and parameters reducing errors from occlusion. It also exploits and expands the pose discriminator based on the kinematic chain space from \cite{wandt_repnet_2019}, using it with temporal data. The authors address the contribution of each module to the final accuracy. However, they do so by adding the modules one at a time to the backbone, which does not provide any insight into whether one module improves or compensates for the errors or performance of another. Further cross-comparison could help determine which module has a greater impact and on which data or context.\\

\cite{liu_attention_2020} Added an attention mechanism to the temporal approach for extracting 3D pose from video. Similarly to \cite{pavllo_3d_2019}, temporal dilated convolution extracts information in the 2D poses sequence. The added attention mechanism selects the frames and tensor outputs that are the most useful for the detection. The temporal attention modules are computed from the distribution of the tensors at each time step. The kernel attention modules are computed from the distribution of the channel outputs of each layer. Both attention modules are propagated within an attention matrix to the following layer. On top of this architecture, multi-scale dilation convolutions (see dilation convolution in the previous described method \cite{pavllo_3d_2019}) with increasing receptive fields are employed to reduce the vanishing gradient issues. This method shows progress over the state-of-the-art methods. An ablation study also shows how attention modules associated with the multi-scale convolution strategy lead to better results, especially on difficult frames (fast motions or partially occluded subjects).\\

The two main contributions of \cite{vedaldi_motion_2020} are: \textit{Motion loss}, a new loss function based on keypoint trajectory and \textit{U-shaped Graph Convolutional Network} (UGCN), a network architecture. Motion loss is based on coordinate vectors. To make it differentiable, so it can be used in a learning architecture, any motion sequence needs to be encoded using a differentiable operator. The authors empirically chose the scalar product (among other tested). The final loss is composed of this motion loss computed with "pairwise motion encoding" and an absolute position reconstruction loss. UGCN uses spatial temporal graph to represent motions \cite{cai_exploiting_2019}. Then, spatial convolutions are applied on each skeleton for each frame before temporal convolutions are applied to the temporal dimension of each joint in the graph. The network architecture is similar to successful ones in  semantic segmentation (e.g. \cite{ronneberger_u-net_2015}. It consists in three stages: downsampling, upsampling and merging. This architecture captures features at different scales, which implies that UGCN explores temporal and spatial information at different scales. This architecture was tested adding upsampling and downsampling  one by one, showing increasing accuracy. The addition of motion loss also drastically improves the results on two benchmarks (Human3.6M and MPI-INF3DHP), improving on the state-of-the-art results.

\subsection{Multi-view}
\begin{table*}[htp]
	\begin{center}
		\begin{tabular}{|l|c|c|c|}
				\hline
				Method & Human3.6M & Total Capture & Input \\
				\hline\hline
				\cite{qiu_cross_2019}&31.17 / 26.21+&29&Multi-view\\			
				\cite{iskakov_learnable_2019} & \textbf{17.7+}/20.80*+ &- &Multi-view\\
				\cite{he_epipolar_2020} &26.9/19.0+&-&Multi-view\\
				\cite{vonPon2016a}& -  & - &Multi-view, IMU\\
				\cite{trumble_total_2017} & 87.3 & 77.0 & Multi-view, Temporal, IMU\\
				\cite{ferrari_recovering_2018} & - & 26.0 &Monocular, IMU\\
				\cite{huang_deepfuse_2019}& 37.5/13.4* & 28.9&Multi-view, IMU\\
				\cite{zhang2020fusing} & - & \textbf{24.6}&Multi-view, IMU\\
				\hline
		\end{tabular}
	\end{center}
	\caption{Accuracy comparison from several state-of-the-art multi-view and multimodal methods. Human3.6M and TotalCapture results reported in absolute MPJPE (lower is better). Techniques with the annotation with + are using extra-training data to obtain the result; the others use the benchmark's recommended protocols. The * annotation indicates results published with procrustes alignment to ground truth poses before evaluation.}
	\label{tab:methods-multimodal}
\end{table*}

\cite{qiu_cross_2019} uses a two-stage prediction process similar to \cite{zhang2020fusing} without taking IMU data as inputs. Instead, they opt to fuse data from multi-view images using projective geometry constraints. This process is done through a convolutional layer that merges pixel data from each view along the epipolar lines using weighted matrices. After this step, fused 2D joints heat maps are generated and 3D pose is inferred through the Pictorial Structure Model with a skeleton model. A recursive variation of this model is used to reduce quantization errors and complexity using a divide and conquer scheme for space discretization. Comparing results from methods providing absolute coordinates, the cross view approach improved the state-of-the-art results at the time and further improves them using additional 2D data during training. It also gives competitive results on the TotalCapture dataset without pre-training and without using the IMU data. It can also be used in different camera setup using pseudo-labeling from 2D pose estimator. The method can be applied to 3D human pose estimation in new contexts without the need of training with 3D ground truths. This method illustrates very well the main approach that consists of using improved results in 2D HPE and translating them into 3D. Here the refinement is done by adding multi-view to improve 2D predictions.\\

\cite{iskakov_learnable_2019} present a learnable way to triangulate human poses. This article proposes two geometric methods for triangulating 3D joint coordinates from multiple view joint heat maps. The first one is an algebraic method based on solving a system of vector equations with 3D coordinates. The second method is a triangulation from volumetric aggregation of re-projections of the 2D heat maps in a voxel grid. Both are weighting the information coming from different views with learnable coefficients. In the volumetric approach, each heat map corresponding to a joint from each different view is sampled into a voxel cube. These volumetric maps (for a specific joints) are then aggregated with a weighting of the impact from different views and fed into a 3D CNN that refines them. The final step is a soft-argmax operation on the resulting 3D heatmaps yielding computable 3D coordinates. Each of these steps is differentiable and the weights of the different convolutional layers at each stage are updated using an absolute loss. The results on Human3.6M for multiple view input are the best ones to date with a 17.7mm MPJPE accuracy (i.e., with the volumetric method and softmax aggregation during the learning stage as described above). On limit of this method is that it needs a correct cropped volume around the human skeleton to work well. Consequently, at least two camera must detect the pelvis joint. The given absolute MPJPE score is also calculated with the removal of several actions due to annotation errors on Human3.6M.Nevertheless, this method reaches the best accuracy for relative pose estimation among all the methods we reviewed.\\

Epipolar Transformers from \cite{he_epipolar_2020} are modules using the attention mechanism as well as knowledge of epipolar geometry. The authors noted that most 2D detection of keypoints did not use 3D features at all: this was the motivation for their "Epipolar Transformer". The goal is to fuse intermediate features from multiple views during 2D inference using projective geometry constraints \cite{qiu_cross_2019}. First, from a detected point in the source view, the module samples all points on the corresponding epipolar line in another view. Then, features across this line are fused according to a computed weighted similarity with the source point. In the end, the obtained feature maps have the same size as the input, which makes the module compatible with any two-stage multi-view system. Any triangulation algorithm can then be applied. On Human3.6M, the authors computed the 3D coordinates with the recursive pictorial structure model of \cite{qiu_cross_2019} using their epipolar transformer: they obtained the best MPJPE score across state-of-the-art, and they did so without using external training data. They also compare their results with pre-training on MPII 2D dataset and get results close to the best methods (19mm against 17.7 from \cite{iskakov_learnable_2019}) with 10\% less parameters and computation operations. The main limitation of the method is that it only works on a fully calibrated multi-camera system, because it needs camera parameters to compute epipolar lines.

\subsection{Multimodal approaches}
\cite{vonPon2016a} propose one of the first method combining multi-view video and IMU data for pose estimation. The authors claim that their method is less intrusive than marker-based MoCap, as only a few sensors are placed on the subjects. They use a representation of human body constraints based on kinematic chains. Using the silhouettes from multi-view extracted with background subtraction and limb orientation from IMU, they minimize a hybrid energy term obtained from orientation and contour consistency with a human mesh model (\cite{loper_smpl_2015}). They provide a thorough analysis of their method (with the TNT15 dataset presented in the same paper and some metrics on HumanEva) showing that video and orientation sensor complement each other. The idea is that IMUs accurately measure joint angles but tend to drift during the experiment, whereas video is better suited to obtain positional information.

\cite{trumble_total_2017} present the first large-scale motion capture dataset that also contains IMU data. The paper describes a 3D pose estimation technique fusing 3D data from multi-view and limb orientation from IMU, while maintaining the temporal context using a Long Short-Term Memory layer over the five past frames. A PVH \cite{grauman_bayesian_2003} is computed from the multi-view video and inputted to a 3D convolutional network with 26 3D joint coordinates as an output. In parallel, kinematic solving provides the same joint coordinates from IMU data. The vectors from both sources are then passed through the LSTM layer and merged into an embedding representing the 3D pose. This way, the authors show that it is possible to learn a "mapping between the predicted joint estimates of the two data sources and the actual joint locations". They evaluate their method on the newly presented TotalCapture dataset. They also evaluate the multi-view part of their pipeline without IMU on Human3.6M.

Following their work with inertial units, \cite{ferrari_recovering_2018} present a monocular method with a moving camera in-the-wild and multiple subjects wearing IMUs. The method uses a 2D multi-person pose estimation algorithm \cite{cao_realtime_2017} and a module that fits the SMPL \cite{loper_smpl_2015} model to IMU data. Then, each 2D skeleton is paired with 3D pose and shape using graph optimization. Next, with the obtained associations model parameters, camera pose and orientation are optimized and fed back for further iterations. When evaluated on the TotalCapture dataset, the method outperforms previous ones by 44mm. This technique allows for the capture of a new dataset in outdoor environments and without marker ground-truths: the 3D Poses in the Wild Dataset.\\

\cite{huang_deepfuse_2019} developed an end-to-end trainable 3D convolutional network witha refinement module based on IMU data. The main idea is to process primitive 3D data from multi-view frames without any transformation. A multi-channel volume, constructed from the segmented human silhouettes and camera parameters, is used as input for the network. 3D voxel confidence heat maps are computed at this stage and can already be used for human pose estimation prediction. The refinement stage merges the volumes constructed from IMU and those constructed from heat maps, as well as the multi-channel volumes. IMU volumes are processed into a "bone cylinder" from quaternion orientation and the previously predicted joint position. Then, all this 3D information is fed into another 3D convolutional network. The architecture uses hourglass networks and residual network modules (see 2D architectures \ref{2DNetworkArchis}) and 3D soft-argmax to extract coordinates. MSE loss is computed at the different levels of the architecture. By randomly stopping information from a camera, the method allows for data augmentation during training, which significantly improves performance on partially captured images. With the vision-only module on a dataset without IMU, the method obtains good overall accuracy. The authors claim that their model can be used in a real-time system because it does not use time sequences. However, they do not provide any performance or speed evaluation. They also point out that their technique does not use a complex human model and is therefore more likely to generalize to new subjects. However, this depends on the performance of the human shape segmentation algorithm at the preprocessing step.\\

Similarly, \cite{zhang2020fusing} fuse IMU with multi-view image data, but this time using a human skeleton model. The approach is to refine 2D joint confidence heat maps with IMU data to then use a part-based model. First, as in \cite{qiu_cross_2019}, they extract 2D joint heat maps. However, the authors use a simple baseline method \cite{xiao2018simple} before merging the heat maps information and the IMU orientation to geometrically estimate the correct coordinates. They call this process Orientation Regularized Network (ORN). This technique can be used with any heat map-based prediction method and they use it to train an end-to-end network that produces more accurate results than state-of-the-art 2D methods not exploiting IMU data. The second part of their work consists of a variant of the Pictorial Structure Model (PSM) (\cite{felzenszwalb_pictorial_2005}, \cite{belagiannis_3d_2014}) which is often used for 3D and 2D HPE. This family of method calculates the most probable human pose within all possible ones in a discrete space. Typically, PSM-type methods uses limb length as the primary constraint. Here, the authors exploit the IMU data to also apply a limb orientation constraint, which improves accuracy. This framework includes a module that can deeply improve results from 2D estimators with inertial data. Note that the 3D part of this system does not use CNN, unlike many other current methods, although it gives excellent accuracy. The authors also use synthetic IMU data to predict 3D pose on the Human3.6M dataset to conclude that their use improves the results by an average of 10mm.\\

Finally, recent works using a single depth sensor \cite{yu_bodyfusion_2017} and \cite{yu_doublefusion_2018} have shown good results for real-time motion capture. The first one uses the iterative closest point algorithm to reconstruct body shapes and the second optimize the SMPL body model. However, these methods are not tested for joint localization error on a pose estimation data set.

\section{Analysis and discussion}
\label{Analysis}
As mentioned in the introduction, this article is structured around the specifications and requirements, which vary according to the fields of application. Today 3D pose estimation is employed in many fields\,:\\

\textit{Human Computer Interface:} There is an increasing number of applications using human pose and gesture to interact with computers. 3D HPE is essential to help robots and machines better understand and respond to human motions.\\

\textit{Security:} The classical applicationis to track people in the indoor and outdoor environment to ensure they do not commit theft or infractions. \\

\textit{Motion Analysis:} This is a broad field that comprises sport and performance analysis, medical study, pose semantics or the study of inter-human interactions.\\

\textit{Entertainment:} Pose estimation can be used for avatar control (e.g., Kinect) or VR and AR refinement and for avatar animation in games and movies.\\

\cite{moeslund_survey_2001} classify them into three categories\,: surveillance, analysis and control. Each category requires different performance regarding accuracy, speed or robustness. However, within the same category, there are variations on these criteria. For example, as the previously mentioned authors state, a control application can be constrained to highly controlled environment (avatar control) or within a generic outdoor scene with varying conditions.\\

For this reason, our review describes each criterion separately and explains in which use case it performs the best. Tables \ref{tab:overall_accuracy}, \ref{tab:overall_robustness} and \ref{tab:overall_speed} classify these methods with accuracy reported in MPJPE. The level of robustness corresponds to the number of assumptions or constraints necessary for correct detection. Lastly, to express the speed, we report whether the model can run in real-time. Each of these tables allow to cross-compare the different methods best suited to the most needed specifications.

\subsection{Accuracy}
\label{Accuracy}
Accuracy is the main criterion used to evaluate markerless human pose estimation methods within the computer vision community. However, it has some limitations that are important to keep in mind. First, most benchmarks compare markerless vision-based methods to results obtained from marker-based optoelectronic systems that are themselves not free from errors. As reported in Section \ref{benchmarks}, some old motion capture datasets contain videos with low resolution and inaccurate annotations.\\

The second limit in accuracy comparison is that, in many cases, it is not the local error rate that is important to the application but the semantics of the pose as whole (See \ref{Metrics}). This can be the case for surveillance systems or human-machine interface for example. When the accuracy requirement (at least with conventional metrics) is low to minimal, see the two other criteria discussed in this review.\\

However, many other fields of research need high accuracy , such as medical science or biology. Typically, the accuracy prerequisite can also be different depending on the analysis. Sometimes, the highest accuracy possible is required regardless of the acquisition and computation procedure complexity (e.g., for research in biomechanics). However, many studies are conducted from human labeled videos with much simpler setups. We  propose a solution based on machine vision for each of these scenarios. 

\begin{table}[t]
	\centering
	\begin{center}
		\resizebox{0.5\textwidth}{!}{\begin{tabular}{c|c|c|c|c}
				\hline
				\multicolumn{5}{c}{Accuracy}\\
				\hline
				Method & Type & Accuracy & Robustness Level &  Real-time \\
				\hline
				\cite{kolotouros_learning_2019}&Monocular&\textbf{41.1}&average&\ding{55}\\
				\cite{wandt_repnet_2019}& Monocular & \underline{50.9} & average & \ding{51}\\
				\cite{iskakov_learnable_2019}&Multi-view&\textbf{17.7}&average&\ding{55}\\
				\cite{he_epipolar_2020}&Multi-view& \underline{26.9}&low& \ding{55}\\
				\cite{cheng20203d}&Temporal&\textbf{40.1}&average&\ding{55}\\
				\cite{vedaldi_motion_2020} & Temporal & \underline{42.6} & average & \ding{55}\\
				\cite{zhang2020fusing}&Multimodal&\textit{24.6*}&low&\ding{55}\\
				\hline
		\end{tabular}}
	\end{center}
	\caption{Best accuracy methods. Methods reported with performance criteria of current real-life applications that use 3D pose estimation for four setups (monocular, temporal, multi-view). Accuracy is reported in MPJPE on Human3.6M dataset bold for best and underlined for second best. The best multi-modal approaches using IMU (marked with *) are evaluated with MPJPE on TotalCapture for comparison.}
	\label{tab:overall_accuracy}
\end{table}

\subsubsection{Highest accuracy methods}
When looking at results of contemporary methods, there is an average gap of 10mm error between monocular and multi-view methods. When multi-view methods are used properly, the combination of geometric knowledge of the scene and learning optimizations can results in errors as low as 20 mm. However, the error is higher in a general context (which is also true for monocular methods). For example, \cite{iskakov_learnable_2019} obtained 34mm MPJPE on a different dataset than the one used for training.\\

IMU sensors are also a good way to improve multi-view detection. However, for now, the results seem close to those obtained with image data alone, with \cite{zhang2020fusing} obtaining the most accurate results. Future evaluations on the TotalCapture dataset \cite{trumble_total_2017} with methods that do not use inertial information could help with the comparison.\\

The strength and weakness of the multi-view methods comes from the fact that they are based on camera parameters. This helps to the generalizability of the system as it can adapt to new camera views \cite{he_epipolar_2020}, but also makes the process more complex as it requires calibration. A solution can be the one suggested by \cite{kocabas_self-supervised_2019}, with a system that computes camera parameters on-the-fly with the detected joint as calibration targets. Multiple images hold much more information than a single one and well-known stereo vision properties are applicable.It turns out that most research has been concentrated on monocular images and 2D estimation, so many other avenues remain unexplored, such as the simultaneous exploitation of temporal and multi-view data.\\

Monocular methods based on video sequences that can be processed offline such as \cite{cheng20203d} and \cite{pavllo_3d_2019} have merit. However, their accuracy is not less than 40 mm MPJPE on Human3.6M, and their ability to generalize is also unproven, as no comparative analysis is yet available for them.

\subsubsection{Simplest but accurate methods}
Monocular methods are the simplest, as they can be fed a simple video input. In many cases, they are a clear improvement over handcrafted annotation, as they have similar accuracy but provide richer 3D information with less input data. When video data is available, it is judicious to favor methods based on temporal data that provide better results (\cite{cheng20203d}; \cite{vedaldi_motion_2020} or \cite{pavllo_3d_2019}). \cite{kolotouros_learning_2019} also propose a method based on single images that is competitive with image sequence techniques (41.1mm MPJPE on Human3.6M). This indicates that their approach of jointly optimizing a mesh model and training a neural network is an effective way to solve monocular pose estimation problems.\\

Although \cite{wandt_repnet_2019} does not achieve the typical 40mm average error, it does so at a low computational cost. Thus, it is a good compromise between accuracy and speed for real-time pose estimation.

In many cases, another important factor is the flexibility of the method. Yet most algorithms were not designed with this in mind, as they are limited to the labels in the training data. For the study of human motion, it is common to have to define custom key points to track limbs, segments or joints. This problem can be solved using DeepLabCut \cite{mathis_deeplabcut_2018} or other methods that are not human-specific \cite{zhang_multiview_2020}. Obviously, these methods still require ground-truth examples for each new key point, but on a much smaller scale. Another possibility is to refine existing models with new custom labels.

\subsubsection{Overall conclusion on accuracy}
The table \ref{tab:overall_accuracy} shows that the most accurate methods have a medium to low level of robustness and that few of them work in real time. The explanation comes from the fact that they are often complex methods that tackle issues such as occlusion with specific modules that increase the overall computational cost. As explained above, the best methods are naturally multi-view techniques that exploit geometric constraints and therefore need calibrations.\\
Currently, the architectures that achieve the best accuracy are two-stage top-down algorithms. They achieve the best results on a variety of benchmarks in monocular image, monocular video, and multi-view configurations. They often build on the success of 2D pose estimation, which is a nearly solved problem (i.e., it achieves average PCK scores above 90\%). Many different approaches are effective: direct regression from 2D to 3D, initialization of human parametric models, exploitation of temporal sequences, occlusion-aware modules, generative models, volumetric input representation, multi-view triangulation, to name the most important (See \ref{Methods} and Fig. \ref{fig:3DMarkerlessSystems}). An interesting point is that, logically, video sequence methods perform better for activities with temporal coherence, such as walking or running, while monocular methods best detect complex static poses better.\\

Another strong distinction can be made between model-based and model-free approaches. While the state of the art in 2D detection does not use human models, 3D detection successfully does. They are either based on Pictorial Structure \cite{belagiannis_3d_2014} or on a 3D mesh model like SMPL \cite{loper_smpl_2015} or SCAPE \cite{anguelov_scape_nodate}. Recent successful propositions also use 2D detection from CNN as initialization for their models. However model-free techniques perform as well or even better in some cases. The detection error for the 3D human pose estimation task has decreased by nearly 70mm over the last decade, mainly due to convolutional networks in different architectures. Interestingly, these improvements were mainly due to better 2D modules, which may indicate that research into the 3D nature of the problem is a fruitful avenue.

\subsection{Robustness}
\label{Robustness}
\begin{table}
	\centering
	\begin{center}
		\resizebox{0.5\textwidth}{!}{\begin{tabular}{c|c|c|c|c}			
				\multicolumn{5}{c}{Robustness}\\
				\hline
				Method & Type & Accuracy & Robustness Level &  Real-time \\
				\hline
				\cite{mehta_xnect_2020}&Monocular&63.6&high&\ding{51}\\
				\cite{xu_denserac_2019}&Monocular&82.4&high&\ding{51}\\
				\cite{hossain_exploiting_2018}&Temporal&58.5&high&\ding{51}\\
				\cite{cheng20203d}&Temporal&40.1&average&\ding{55}\\
				\cite{liu_attention_2020}&Temporal&45.1&average&\ding{55}\\
				\cite{iskakov_learnable_2019}&Multi-view&17.7&average&\ding{55}\\
				\cite{ferrari_recovering_2018}&Multimodal&26.0*&high&\ding{55}\\
				\hline
		\end{tabular}}
	\end{center}
\caption{Most robust methods. Robustness level is defined as follows: 1 - 3 assumptions: high level, 3 - 4 assumptions: average level and 5 or more: low level.}
		\label{tab:overall_robustness}
\end{table}
Robustness is usually assessed through changes in the accuracy during cross-dataset evaluation. \cite{moeslund_survey_2001} propose to express robustness as the number of assumptions required for a motion capture configuration to be operational. They define twenty assumptions relative to the acquisition protocol or environment and the subject’s appearance. Some of them have been overcome by all current method (subject wearing specific clothes or static monochrome backgrounds), but others are still very much debated (occlusions, single person or no camera motion). Below are assumptions that remain in the state of the art:

\textit{Motion Constraints\,:} No camera motion, no fast motion of the subject, no occlusion (no severe occlusion, no auto-occlusion). For temporal methods, the number of frames and the acquisition frequency may also be a limitation.

\textit{Environment Constraints:} Extra hardware (IMU, Laser Scans), Multiple cameras (with or without camera parameters), etc.

\textit{Subject Constraints: } A single person, a known first pose, a motion parallel to the camera plane (for a model-based approach)

\begin{table}[t]
	\centering
	\begin{center}
		\begin{tabular}{c|c|c}
				\hline
				Method & \#Frames & Causal\\
				\hline
				\cite{hossain_exploiting_2018}&5&\ding{51}\\
				\cite{cai_exploiting_2019}&7& \ding{55}\\
				\cite{pavllo_3d_2019}&243&\ding{51}\\
				\cite{cheng_occlusion-aware_2019}&256&\ding{55}\\
				\cite{cheng20203d}&128&\ding{55}\\
				\cite{liu_attention_2020}&243&\ding{51}\\
				\cite{vedaldi_motion_2020}&96&\ding{55}\\
				\hline
				
		\end{tabular}
	\end{center}
	\caption{Monocular temporal models assumptions. Needed number of frames to obtain the optimal accuracy and whether the system can be adapted to only use past frames (for real-time use)}
	\label{tab:temporal_assumption}
\end{table}

\subsubsection{Background, Lighting and Clothes}
Most appearance constraints are no longer required, because convolutional networks do much better at identifying meaningful visual invariants than former hand-crafted feature extractors. Yet, it is difficult to evaluate generalization to in-the-wild situations, mainly because large motion capture datasets are still captured in indoor studios. Data augmentation can provide new scenes with background extraction or change the lighting. With the release of new commercial markerless solutions, new benchmarks such as MPI-INF-3DHP \cite{mehta_161109813_2016} also include real outdoor data.

\subsubsection{Special Hardware \& Calibrated Systems}
Two constraints are still relevant: first, the need for specific hardware, second, the need for calibration in multi-view configurations. The two most commonly used extra hardware added for pose estimation are inertial motion units and depth sensors such as infrared or time-of-flight cameras. Inertial motion unit provide additional information on the limbs orientation but suffer from drift in their results after a short usage. They are also more practical than reflecting markers and motion capture suits but are still intrusive for the subject. Different depth sensors have also been used to infer the depth of joints directly or to construct more complex features as a pre-processing step of pose estimation. Less frequently some methods use the 3D scan of each subjects that is captured to fit shapes parameter of human body models. Logically, while most research is conducted on monocular methods, they are always outperformed by 10 to 20 mm MPJPE by multi-view techniques. In multi-view systems, the calibration step is a frequent requirement. Multi-view methods that do not use or use partial calibration need more views to achieve acceptable accuracy, while others can produce good results with fewer cameras but need extrinsic parameters (see table \ref{tab:multiview_assumptions}).

\begin{table}[t]
	\centering
	\begin{center}
		\resizebox{0.5\textwidth}{!}{\begin{tabular}{c|c|c|c}
				\hline
				Method & Calibration & Additional & \#Views  \\
				&&Hardware& $<$ 40 MPJPE\\
				\hline
				\cite{trumble_total_2017}&\checkmark&\checkmark&4\\
				\cite{qiu_cross_2019}&\checkmark&&4\\
				\cite{kocabas_self-supervised_2019}&&&4\\		
				\cite{iskakov_learnable_2019} &\checkmark&&2\\
				\cite{he_epipolar_2020} &\checkmark&&3\\
				\cite{huang_deepfuse_2019}&\checkmark&\checkmark&4(8)\\
				\cite{zhang2020fusing} &\checkmark&\checkmark&4(8)\\
				\hline
				
		\end{tabular}}
	\end{center}
	\caption{Multi-view models hardware-related assumptions. The number of camera view used to achieve less than the baseline 40mm MPJPE error (best results from monocular methods) on Human3.6M is also shown (and on TotalCapture under parenthesis).}
	\label{tab:multiview_assumptions}
\end{table}

\subsubsection{Single Person vs Multi-Person}
The strong assumptions that are still used for many 3D human pose estimation frameworks are related to camera or subject motion and acquisition protocol. One of them is the limitation to one person in the image. As top-down approaches are the most popular, many methods assume or even take as an input a single person. For this reason, some authors recommend the use of off-the-shelf person detectors to crop to the area of the image containing the individual subjects. In this way, the pose estimation algorithms can be applied to each area individually. However, this idea needs to be adapted in multi-view and temporal settings to track each different subject. The main limitation is when subject parts are overlapping in the image, hence indistinguishable by the person detectors. Some research addressed this issue in 2D, but it is still an open problem for 3D with few specific methods \cite{mehta_single-shot_2018} \cite{mehta_xnect_2020}.

\subsubsection{Motion Restriction}
Former markerless motion tracking systems were sometimes constrained to slow motion of only few limbs to perform good detection. It is less and less the case, but there is still some difficulty in predicting quick motions (e.g., in sports video). \cite{cheng20203d} suggest that temporal and spatial data at different resolutions can be a solution to this issue. A new assumption that can be added for methods based on temporal data\,: if the video footage is not long enough to provide sufficient information, this can be an issue for real-time inference and even for accuracy. Additionally, temporal methods often use information in the future frame, which is not suitable for real-time. Table \ref{tab:temporal_assumption} show that these methods perform best with varying temporal receptive fields. Some methods only need a few past and future images, while the accuracy of others saturates at more than 200. These methods can be adapted to shorter video clips and to real-time application using only past frames, but at the price of a decrease in accuracy. Another strong constraint is the use of video from moving cameras, but this is rarely addressed \cite{ferrari_recovering_2018}. Many applications can work with the assumption of fixed cameras, but there is a substantial amount of video data produced with moving camera coordinate systems (e.g., in outdoor sports).

\subsubsection{Occlusions}
Recent solutions predict poses even in the presence of small occlusion, but this remains one of the main challenges in a monocular approach. In most real-world or multi-person scenarios, this needs to be addressed. To this end, many optimizations at training time are employed: data augmentation or occlusion-specific modules (\cite{cheng20203d}; \cite{cheng_occlusion-aware_2019}; \cite{huang_deepfuse_2019}). Another solution could be to consider the 3D scene around the subject. \cite{hassan_resolving_2019} show that combining 3D reconstruction of indoor scene and volumetric models of the human body can help overcome the occlusion issues. Their "Proximal Relationships with Object eXclusion" method enforces physical constraints such as contacts with a static environment.

\subsubsection{Generalization}
Although neural network models are data-oriented algorithms, few analyses are performed on generalization to new contexts and in-the-wild situations. Protocols for benchmarks address generalization to different subjects, but the background scene and the actions rarely change. Multi-view models generalize best, with good performance in cross-validation between data sets, likely due to the geometric information provided by the camera projection matrices they often use. Yet, new benchmarks are necessary to properly assess generalization. Future research will likely provide carefully designed naturalistic datasets with new labeling solutions, or datasets augmented with new image processing techniques, and possibly benchmarks composed of synthetic poses, to challenge future methods.

\subsubsection{Overall Conclusion on Robustness}

Table \ref{tab:overall_robustness} shows the less constrained methods and their performance. Robustness relates to the number of assumptions (the fewer the better). For monocular techniques, the multi-person methods trained on complex datasets containing severe occlusions logically impose fewer constraints. For temporal techniques, the ones that do not need future frames for inference perform best. \cite{hossain_exploiting_2018} achieves maximum accuracy with the fewest images \cite{liu_attention_2020}. \cite{cheng203d} address fast motion and occlusions but require a higher acquisition frequency and a wider temporal receptive field to produce better results. The most robust multi-view method is \cite{iskakov_learnable_2019} because it does not require any special hardware and can work with only two cameras while achieving acceptable accuracy (see table \ref{tab:multiview_assumptions}). Finally, \cite{ferrari_recovering_2018} can perform multi-person pose estimation in the wild with mobile cameras. Yet, the low constraints in terms of environment and subject come at the cost of high constraints in terms of additional hardware (IMU and scans).

Throughout this literature review, we have found that achieving the highest accuracy is the primary concern of novel methods. We also found that system robustness is regularly overlooked when evaluating algorithms and that the complexity of the novel technique is rarely considered, especially for fully engineered systems. Our impression is that the general emphasis on highest accuracy is somewhat misleading, and that the search for "good enough" systems depending on the application domain might be at least as important. For example, very high accuracy is mostly irrelevant for surveillance, but robustness and real time operation matters. For healthcare professionals monitoring patient recovery, robustness is paramount, provided that accuracy is enough (i.e., comparable to that of the human eye) but real-time operation only desirable.\\

\subsection{Speed}
\label{Speed}
\begin{table}
	\centering
	\begin{center}
		\resizebox{0.5\textwidth}{!}{\begin{tabular}{c|c|c|c|c}			
			\multicolumn{5}{c}{Speed}\\
			\hline
			Method & Accuracy & Robustness Level & Speed & GPU used \\
			\hline
			\cite{mehta_vnect:_2017}&80.5&average&30fps&N/A\\
			\cite{wandt_repnet_2019}&50.9&average&20 000fps&Nvidia Titan X\\
			\cite{xu_denserac_2019}&82.4&high&120fps&N/A\\
			\cite{mathis_deeplabcut_2018}& - &average&30fps&Nvidia 1080Ti\\
			\cite{mehta_xnect_2020}&63.6&high&30fps&N/A\\
			\cite{hossain_exploiting_2018}&58.5&high&300fps&Nvidia Titan X\\
			\cite{liu_attention_2020}&45.1&average&3000fps&Nvidia Titan RTX\\
			\cite{pavllo_3d_2019}&46.8&average&150 000fps&Nvidia GP100\\
			\cite{trumble_total_2017}&87.3&low&25fps&N/A\\
			\cite{huang_deepfuse_2019}&37.5&low&25fps&Nvidia 1080Ti\\
		\end{tabular}}
	\end{center}
	\caption{Real-time methods. Along with the other criteria for comparison, the speed (in frames inferred per second) and the graphical card used are reported. Note that methods such as \cite{hossain_exploiting_2018} and \cite{wandt_repnet_2019} are not considering the speed of the 2D detector in the first stage of their techniques when reporting fps.}
	\label{tab:overall_speed}
\end{table}

It is easier to evaluate the performance in terms of speed by simply observing the complexity or the number of single operations needed for any one method (using floating-point operation \textit{"FLOPs"}). When possible, knowing at what frame rate an inference can be made is also interesting for real-time application such as monitoring, surveillance or virtual reality. In these cases, the main constraint is the whole system latency, which should not exceed specified limits.

\subsubsection{Real-time}
Not all methods are complete motion capture systems such as \cite{huang_deepfuse_2019} or \cite{mehta_vnect:_2017} that report 25 and 30 frames per second for detection. Some only provide the inference timeper frame, without testing in a real life acquisition scenario such as \cite{hossain_exploiting_2018} and \cite{martinez_simple_2017} who each report 3ms per frame. The deepest architectures are often not suitable for real-time applications, as a forward pass in the network takes too long. To gain insight into the complexity of these models, one can look at the number of parameters that they learn.

\subsubsection{Training Time}
Most of the reviewed methods can be trained to adapt to new challenging contexts or refined on new data. It is also important to have an idea of the training time when an application might consider online training. The number of parameters is a good  indication of the depth of the architecture (see Table \ref{tab:parameters}), which can also be calculated using backbone 2D detection networks in many cases.

\begin{table}[ht]
	\begin{center}
		\begin{tabular}{c|c}
			\hline
			Method & \#Parameters\\
			\hline
			\cite{martinez_simple_2017}& 4-5M\\			
			\cite{sun_integral_2018} HG/Res50/Res101&26M/26M/45M\\			
			\cite{pavllo_3d_2019}&16.95M\\
			\cite{qiu_cross_2019}&560M\\
			\cite{iskakov_learnable_2019} alg.&80M\\
			\cite{iskakov_learnable_2019} vol.&81M\\
			\cite{he_epipolar_2020}&69M\\
			\hline
		\end{tabular}
	\end{center}
	\caption{Number of reported parameters of different reviewed techniques.}
	\label{tab:parameters}
\end{table}

\subsubsection{Overall Conclusion on Speed}
In the speed part of table \ref{tab:overall_speed}, we report all methods that can run in real time. The inference speed is in frames per second and the GPU used is also specified. Robustness is variable in this collection of methods, and the accuracy seems a bit below average for the fastest methods. The most accurate is \cite{huang_deepfuse_2019}, which achieves 37.5 MPJPE on Human3.6M without using its IMU component (it also performs well on TotalCapture with the addition of IMU data).

\subsection{Recommendations for users}
After this analysis according to the performance criteria, here are our suggestions for the different types of application\,:\\

\textit{Human Computer Interface}: for this category of applications, the priority is good accuracy and real-time performance. Robustness depends on whether the system operates in a pre-defined environment or "in-the-wild". With these specifications, temporal methods that can run in real-time such as \cite{pavllo_3d_2019} or \cite{liu_attention_2020} correspond best. \cite{hossain_exploiting_2018} requires fewer frames to be computed but its accuracy is lower. Other more robust methods could be multi-person ones, such as \cite{mehta_xnect_2020}, but control applications usually only interact with one subject. Finally, \cite{huang_deepfuse_2019} is the most accurate method running in real-time, but it requires multiple views to become robust.\\

\textit{Security}: Traditional monitoring systems generally need to be more robust to operate in varied and changing real-world environments and subjects. Speed is also important because infractions need to be identified in real-time. Finally, the higher precision is less important because it is the whole semantics of the movement that is important to identify the subject's actions. If you are looking for methods running in real-time with the highest robustness, consider monocular (\cite{xu_denserac_2019}) and temporal (\cite{hossain_exploiting_2018}) methods. The methods proposed by \cite{mehta_xnect_2020} allows for an even greater robustness with multi-person detection in real-time, if necessary. Less robust methods can also be considered for higher accuracy (e.g. \cite{pavllo_3d_2019} or \cite{liu_attention_2020}) or speed (e.g. \cite{wandt_repnet_2019}).\\

\textit{Motion Analysis}: these applications typically run in lab-controlled environment and computation is performed offline. Accuracy is the most important criteria with less consideration for real-time or high robustness. However, human performance captured in "real-life" scenarios and the need of less intrusive techniques (e.g., medical diagnosis or rehabilitation) demands better robustness than classical marker-based methods (\cite{colyer_review_2018}). Multi-view configurations now produce results with an average error per joint of less than 30 mm (\cite{iskakov_learnable_2019}, \cite{he_deep_2015}). When multiple cameras or calibration are not possible, monocular  temporal methods such as \cite{kolotouros_learning_2019} or \cite{cheng20203d} can be considered. Finally, if simplicity and usability are required, easy-to-use and flexible systems based on good 2D detectors such as \cite{mathis_deep_2019} or \cite{martinez_simple_2017} produce efficient results.\\

\textit{Entertainment}: Motion capture for animation or VR is usually done in a controlled environment, so robustness is not the main concern. Accuracy is important because the generated poses and movements must be realistic (see \ref{Metrics} for structural and perceptual metrics). Finally, speed is less important for offline processing that generates avatar or animated characters, whereas real-time is needed for controls in video games or VR. In the first case, multi-view markerless systems can replace classical motion capture systems at a lower cost (e.g. \cite{iskakov_learnable_2019} or \cite{he_epipolar_2020}). IIn the second case, real-time operation is possible with \cite{huang_deepfuse_2019}, which also provides high accuracy in multi-view configurations (and even higher if adding IMUs to the subject is feasible).

\subsection{Challenge for future research}
Many problems concerning the estimation of the human pose in 3D have still not been solved.  The most discussed is that the accuracy is still insufficient for some applications (e.g., in motion analysis). Currently, many different monocular and temporal approaches achieve an average joint position error of about 40 mm, while multi-view approaches achieve about 20-30 mm. This leads to important challenges:

For \textit{monocular} techniques, the obstacle to their widespread use is the consistency of their detection over space and time. Difference in accuracy among joints needs to be addressed and temporal consistency (reducing jitter) of motions needs to be better enforced. Reaching 90\% of joints accurately detected, as 2D pose estimators do, is a goal for the coming years (the average 3DPCK reaches about 85\% with evaluations on MPI-INF-3DHP).

For \textit{multi-view} setups, the issue is that they present results close to those of the best monocular methods despite the access to epipolar geometry and 3D information. The combination of recent advances in deep learning (e.g., recurrent network, attention, etc.) and strong prior information about scene structures from camera calibration could take this a step further.  Another avenue for applications in more controlled environments could be the use of other modalities such as IMU or depth sensors. Multi-modal information fusion could then lead to even better results.

\textit{Real-time} performances are achieved by many methods, but using powerful graphical cards. Porting real-time estimation to the average commercial equipment remains a challenge. Currently, most proposals rely on multi-stage computations, but future research could draw on single-shot object detectors (e.g. \cite{liu_ssd_2016} (SSD), \cite{redmon_you_2016} (YOLO)) to produce reasonable results faster. Similarly, research on real-time 3D human pose estimation with multi-view cameras is developing. Virtual reality could benefit from such systems for gesture-based control.

Finally, there are strong assumptions about occlusions and multi-person configurations that do not yet allow pose estimation to be applied to any video or image. Complex in-the-wild data sets, sometimes including difficult poses (generated or captured by motion), are beginning to emerge, which suggests that these are promising research questions. 

\section{Conclusion}
Human pose estimation has been one of the focal points of computer vision in recent years. Deep learning has improved the results by a significant amount. However, the most accurate techniques use various architectures (temporal convolutional networks, 3D human body models or learnable triangulation) depending on the input data (single images, videos sequence or multi-view images). What these methods have in common is the use of 2D detection as an intermediate step. However, the diversity of approaches in the new state-of-the art methods demonstrates that a consensus has not yet been established.\\

Improved results in the near future will require richer datasets, computational parsimony, and ease of use. Access to new and richer datasets, including a wide variety of poses, movements, and contexts, is essential for robustness. This could be facilitated by new learning processes working with partially annotated data based on monocular videos, but also by commercial markerless tracking tools, which could feed these repositories with better images in "natural conditions", without markers and in an outdoor environment. Parsimony in terms of computational cost paves the way for real-time operation, less expensive hardware architectures, and a welcome reduction in power consumption. Ease of use, as was the case with DeepLabCut, is a key to widespread adoption, which in turn challenges the algorithms to solve new questions and use cases.\\

Finally, we believe that important future developments could come from new ways of approaching the full richness of human pose estimation, by reasoning directly on the 3D nature of the problem (using voxel map representations, multi-channel volumes...), by employing less constrained multi-view approaches that can function with fewer cameras, or by transposing temporal methods (transformers architecture or temporal convolutional networks) from the monocular to the multi-view.

\section*{Acknowledgments}
This project was supported by the LabEx NUMEV (ANR-10-LABX-0020) within the I-SITE MUSE.
This research was partially supported by the HUT project co-financed by the European Regional Development Fund (ERDF) and the Occitanie Region.
The funders had no role in study design, data collection and analysis, decision to publish, or preparation of the manuscript.

\nocite{*}
\bibliographystyle{model2-names}
\bibliography{ref}
\end{document}